\documentclass[preprint,5p]{elsarticle}

\usepackage{morefloats}
\usepackage[linesnumbered,ruled,vlined]{algorithm2e}
\usepackage{graphicx}

\usepackage{multirow}
\usepackage{booktabs}
\usepackage{amssymb}
\usepackage{amsmath}
\usepackage{subfig}
\usepackage{float}
\usepackage{graphicx}
\usepackage{epstopdf}
 \usepackage{natbib}

\usepackage[figuresright]{rotating}
\usepackage{caption}

\usepackage{rotating, blindtext}
\usepackage[dvipsnames]{xcolor}

\usepackage{framed}
\usepackage{balance}

\makeatletter
\newcommand{\removelatexerror}{\let\@latex@error\@gobble}
\makeatother

\begin{document}

\begin{frontmatter}

\title{Golden Tortoise Beetle Optimizer: A Novel Nature-Inspired Meta-heuristic Algorithm for Engineering Problems }




\author[rvt]{Omid Tarkhaneh\corref{cor1}}
\ead{ tarkhanehomid@gmail.com}

\author[rvts]{Neda Alipour}
\ead{nedalipur@gmail.com }
\author [sa]{Amirahmad Chapnevis}

\ead{chapnevisa@vcu.edu }
\author[rvtc]{Haifeng Shen}
\ead{haifeng.shen@acu.edu.au}

\cortext[cor1]{Corresponding author.}
\address[rvt]{Department of Computer Science, University of Tabriz, Tabriz, Iran.}
\address[rvts]{Deparment of Electrical Engineering, Shahed University, Tehran, Iran}
\address[sa]{Department of Computer Science, Virginia Commonwealth University, Richmond, USA}
\address[rvtc]{Peter Faber Business School, Australian Catholic University, Sydney, Australia}

\begin{abstract}
This paper proposes a novel nature-inspired meta-heuristic algorithm called the Golden Tortoise Beetle Optimizer (GTBO) to solve optimization problems. It mimics golden tortoise beetle's behavior of changing colors to attract opposite sex for mating and its protective strategy that uses a kind of anal fork to deter predators. The algorithm is modeled based on the beetle's dual attractiveness and survival strategy to generate new solutions for optimization problems. To measure its performance, the proposed GTBO is compared with five other nature-inspired evolutionary algorithms on 24 well-known benchmark functions investigating the trade-off between exploration and exploitation, local optima avoidance, and convergence towards the global optima is statistically significant. We particularly applied GTBO to two well-known engineering problems including the welded beam design problem and the gear train design problem. The results demonstrate that the new algorithm is more efficient than the five baseline algorithms for both problems. A sensitivity analysis is also performed to reveal different impacts of the algorithm's key control parameters and operators on GTBO's performance.
\end{abstract}

\begin{keyword}
Golden Tortoise Beetle; optimization; Meta-heuristic; sensitivity analysis.
\end{keyword}

\end{frontmatter}


\section{Introduction} 
\label{sec:intro}
Optimization is a common problem in many fields ranging from engineering design to holiday planning~\cite{yang2010nature}. In these activities, the aim is to optimize an objective such as profit or time, or a combination of multiple objectives. An optimization process consists of multiple input variables, a defined objective function, and an output which is the cost or the fitness of the objective function~\cite{yang2010nature, rajabioun2011cuckoo}. Different algorithms have been proposed to solve different optimization problems and based on the nature of these algorithms, they can be classified into three main categories: deterministic algorithms, stochastic algorithms, and hybrid algorithms. A deterministic algorithm has no randomness and it follows a single path to reach the goal. The hill-climbing algorithm~\cite{yang2010nature, johnson2002class} is one such example of this category. Stochastic methods are based on the concept of randomness, which helps them try different ways of searching. One typical example of this type is the Genetic Algorithm (GA) which utilizes semi-randomness when searching into a problem area~\cite{yang2010nature, holland1992genetic}. In other words, such an algorithm tries to look for a solution using a different path in each separate run. A hybrid algorithm is the composition of stochastic and deterministic methods and one basic example is the hybrid of GA operators and the hill-climbing method. For example, the starting point in the deterministic method can be generated by random mechanisms~\cite{yang2010nature}.

Meta-heuristics, as a special type of stochastic algorithms, have been widely adopted in solving different optimization problems, including but not limited to, image segmentation~\cite{tarkhaneh2019adaptive}, data clustering~\cite{hatamlou2013black}, or feature selction~\cite{zhang2020binary, mafarja2017hybrid}. Simplicity, flexibility, derivation-free mechanism, and local optima avoidance are some of the advantages of these algorithms~\cite{mirjalili2014grey, mirjalili2015ant}. First, simplicity is inherited from natural concepts, which assist scientists to understand meta-heuristic algorithms and apply them to solve real problems. Second, flexibility makes these algorithms applicable to a wide range of problems without any change to their internal structures as they can be used as blackboxes~\cite{mirjalili2014grey, mirjalili2015ant}. Third, meta-heuristic algorithms have established the so-called derivation-independent mechanism where an optimization process starts with random solutions without calculating the derivative of search spaces to seek the optima, making them suitable for unknown spaces. Last, these algorithms are capable of avoiding local optima due to the random operators that stochastic meta-heuristics such as evolutionary algorithms utilize. Most meta-heuristic algorithms use a two-phase search process: diversification (exploration) followed by intensification (exploitation)~\cite{gandomi2013cuckoo}. In the diversification phase, an optimizer utilizes stochastic operators to randomly and globally investigate the promising area(s) of the search space. In the intensification, it focuses on local searching around the promising area(s) derived from the previous phase~\cite{olorunda2008measuring, alba2005exploration, lin2009auto}.

Nature-inspired optimizers are a specific type of meta-heuristic algorithms that are typically modeled after biological behaviors, physical phenomena, or evolutionary concepts. These algorithms have a wide spectrum of applications ranging from engineering optimization~\cite{cagnina2008solving, yang2012bat} to deep learning~\cite{papa2017quaternion}, and to coordinating robots~\cite{de2018swarm}. Section~\ref{lab:relwork} details some of the representative nature-inspired meta-heuristic algorithms. Due to the stochastic nature of an optimization process, finding a suitable balance between diversification and intensification is one of the most challenging tasks in the development of a meta-heuristic algorithm. Some nature-inspired algorithms suffer from performance degrade when they are applied to composition engineering problems such as the welded beam design problem~\cite{mirjalili2016multi} or the gear train design problem~\cite{mirjalili2015ant} due to the difficulty in striking a balance between exploration and exploitation. This motivated us to search for a better solution because according to the No Free Lunch (NFL) theorem~\cite{wolpert1997no}, any two optimization algorithms perform equally when solving all problems. In other words, one algorithm may be superior in solving one set of problems but ineffective in solving another set.

In this paper, we present a novel nature-inspired optimization algorithm called the Golden Tortoise Beetle Optimizer (GTBO) that is inspired by the nature-life and the behavior of golden tortoise beetle. In this algorithm, special operators are mathematically modeled after the mating behavior and the survival strategy of golden tortoise beetle using heuristic techniques to generate solutions for solving optimization problems. To assess the performance of the proposed algorithm against baselines, we employed 24 well-known benchmark functions including unimodal, multimodal, and composite functions~\cite{digalakis2001benchmarking, molga2005test, suganthan2005problem} to compare and contrast their exploration, exploitation, local minima avoidance, and convergence rates. We also conducted Wilcoxon rank-sum test to prove that the GTBO's superior performance is statistically significant~\cite{wilcoxon1992individual}. As the employed benchmark functions are typically unconstrained or with only simple constraints while real-world problems often have specific constraints and contain unknown search areas~\cite{yang2020nature}, we applied the proposed algorithm to two well-known real engineering problems including the welded beam design problem and the gear train design problem~\cite{mirjalili2015ant} and the results again confirmed GTBO outperformed the baseline algorithms. We further conducted a sensitivity analysis to reveal different impacts of the algorithm's key control parameters and operators on its performance for the two engineering problems.

The main contributions of the paper are summarized as follows:
\begin{enumerate}
\item We propose a novel nature-inspired algorithm for solving optimization problems that are unprecedentedly modeled after golden tortoise beetle.
\item The proposed algorithm creates a decent search mechanism that is underpinned by the two unique color-switching and survival operators.
\item The proposed algorithm is efficient for real-world engineering problems and problems with unknown search areas.
\item The proposed algorithm has few control parameters to adjust and is easy to implement.
\end{enumerate}

The rest of the paper is structured as follows: Section~\ref{lab:relwork} presents a literature review of nature-inspired meta-heuristic algorithms. Section~\ref{lab:fundamentals} describes the proposed GTBO algorithm. Experiments setup and results are presented in Sections~\ref{lab:experiments} and~\ref{lab:results} respectively, followed by the applications of GTBO in two engineering problems in Section~\ref{se:engopt}. Finally, Section~\ref{lab:conclusion} concludes the work and suggests some directions for future research.

\section{Related Work}
\label{lab:relwork}
Nature-inspired meta-heuristic optimization algorithms can generally be categorized into evolutionary, physics-based, and swarm intelligence (SI) algorithms~\cite{mirjalili2016whale}.

\subsection{Evolutionary algorithms}
Evolutionary algorithms (EAs) are typically inspired by evolutionary concepts from nature for solving optimization problems. One of the well-known algorithms is the Genetic Algorithm (GA). This algorithm is related to Darwin's theory of evolutionary concepts and was proposed by Holland~\cite{holland1992genetic} in 1992. This simple algorithm starts the optimization process by generating a set of initial random solutions as candidate solutions (individuals) and the whole set of solutions is considered as a population. Each new population is created by the combination and mutation of individuals in the previous generations. Individuals with the highest probability are randomly selected for creating the new population.
There are some other studies based on EA such as Differential Evolution (DE)~\cite{storn1997differential}, Evolutionary Programming (EP)~\cite{yao1999evolutionary}, Evolution Strategy (ES) ~\cite{sprave1994linear}, Genetic Programming (GP)~\cite{koza1992genetic}, and Biogeography-Based optimizer (BBO)~\cite{simon2008biogeography}.

The proposed Golden Tortoise Beetle Optimizer (GTBO) is also an evolutionary algorithm. To the best of our knowledge, there has been no evolutionary technique in the literature that imitates the behavior of golden tortoise beetle.

\subsection{Physics-based algorithms}
These algorithms are developed by mimicking some of the physical rules, such as gravity, electromagnetic force, and weights. One representative algorithm is the Gravitational Search Algorithm (GSA) ~\cite{rashedi2009gsa}, which was inspired by Newton's law of gravity and the law of motion. This algorithm starts the searching process by utilizing a random collection of agents that have mass interaction with each other. Throughout iterations, the masses are attracted to each other according to the gravitational forces between them. The heavier masses attract bigger force. As such, the global optimum will be found by the heaviest mass, while other masses will be attracted according to their distances. GSA can produce high-quality solutions, but it has some drawbacks, such as slow convergence and the tendency to become trapped in local minima. Some of the other well-known physics-based algorithms include Charged System Search (CSS)~\cite{kaveh2010novel}, Central Force Optimization (CFO)~\cite{formato2007central}, Artificial Chemical Reaction Optimization Algorithm (ACROA)~\cite{alatas2011acroa}, Black Hole (BH) algorithm~\cite{hatamlou2013black}, Ray Optimization (RO) algorithm~\cite{kaveh2012new}, and Galaxy-based Search Algorithm (GbSA)~\cite{shah2011principal}.

\subsection{Swarm Intelligence (SI) algorithms}
SI algorithms are inspired by the population-based behavior of social insects as well as other animal species. For example, Particle Swarm Optimization (PSO) algorithm~\cite{carvalho2007particle} models the swarming behavior of a herd of animals and flocks of birds. This algorithm defines individuals as particles and tries to find the optima by utilizing information obtained from each particle. This algorithm, similar to the GA method, starts the optimization process with a set of initial populations, or in fact random solutions as candidate solutions, which are distributed throughout the search space. Based on the best solution, the position and velocity of each particle are updated over the subsequent iterations. In addition, there is another set called velocity which preserves the number of moving particles that have moved with the best solution gained by the swarm. The PSO algorithm has fewer parameters for tuning and is good for multi-objective problems. The most important disadvantages of PSO are that they easily fall into local optimum in high-dimensional spaces and that they have a low convergence rate in iterative processes.

Another well-known SI algorithm is Ant Colony Optimization (ACO)~\cite{dorigo2006ant} that was inspired by the social behavior of ants in nature. The goal of this algorithm is to find the best and shortest path to the source of food. Candidate solutions are improved over the course of an iteration. ACO algorithm can search among a population in parallel and also allow the rapid discovery of good solutions. However, it has an uncertain time to convergence as well as dependent sequences of random decisions. A further novel algorithm is Grey Wolf Optimization (GWO)~\cite{mirjalili2014grey}, which was inspired by the leadership hierarchy and hunting mechanism of grey wolves. There are four levels of wolves called alpha, beta, omega, and delta respectively each of which has a certain responsibility. The hunting process can be divided into three phases, including tracking and chasing the prey, harassing the prey, and finally attacking the prey. However, this algorithm sometimes has a low local search ability as well as a slow convergence. In addition, there are a lot of other SI algorithms such as Artificial Bee Colony Algorithm (ABC) ~\cite{karaboga2007powerful}, Black Widow Optimization (BWO)~\cite{hayyolalam2020black}, Cuckoo Search Algorithm (CS)~\cite{yang2009cuckoo}, Ant Lion Optimizer (ALO)~\cite{mirjalili2015ant}, Artificial Fish-Swarm Algorithm (AFSA)~\cite{li2003new}, Wasp Swarm Algorithm~\cite{pinto2007wasp}, and Fruit fly Optimization Algorithm (FOA)~\cite{pan2012new}.

Table~\ref{tab:Methods} briefly compares some of the nature-inspired meta-heuristic optimization algorithms including the proposed new algorithm GTBO in terms of convergence, local minima avoidance, exploration, exploitation, and balance between exploration and exploitation. Regarding the GA, there is a common problem of local minima stagnation which is due to the lack of population diversity. Another issue is the poor balance between exploration and exploitation, which can be solved by adjusting the selection approach~\cite{hussain2019trade}. As for PSO, a major shortcoming is also to do with a poor balance between exploration and exploitation. Besides, it cannot properly maintain particle diversity, which results in premature convergence. However, this algorithm has a high exploitation capability and is easy to implement~\cite{arani2013improved}. For the ABC algorithm, it can get trapped into local minima especially in solving complex multimodal problems and it has poor exploitation. However, the algorithm benefits from good exploration~\cite{gao2012global, karaboga2009comparative}. Similarly in algorithms such as GSA, there is a risk of premature convergence and the balance between exploration and exploitation is also poor. GSA uses complex operators that require high computational time and hence has a slow searching speed especially in the last few iterations so it cannot guarantee global optima in some cases~\cite{yadav2018harmony}. Other algorithms such as ALO~\cite{mirjalili2015ant} and BWO~\cite{hayyolalam2020black} have good exploitation, but their exploration is poor especially in multimodal and composite functions.

In contrast, the proposed GTBO performs all well on convergence rate, good balance between exploration and exploitation, and local minima avoidance. With its unique operators, it can effectively explore and exploit the search space, which in turn ensures a good convergence speed and maintains a good balance between exploration and exploitation. Furthermore, random selection of some solutions helps the algorithm increase the probability of avoiding local minima.

\begin{table*}[tp!]
\fontsize{8pt}{12pt}
\selectfont

\centering
\caption{Comparison of optimization algorithms}
\label{tab:Methods}
\begin{tabular}{ lllllll }
\hline
Algorithm & Convergence &Local Minima Avoidance & Exploration & Exploitation & Balance of Exploration \& Exploitation \\ \hline
\multirow{2}{*}{} GA & Poor & Poor & Good & Average & Poor \\

\multirow{2}{*}{} DE & Average & Average & Good & Average & Poor \\

\multirow{2}{*}{} GSA & Poor & Poor & Good & Average & Poor \\

\multirow{2}{*}{} ACO & Average & Good & Poor & Good & Average \\

\multirow{2}{*}{} PSO & Poor & Poor & Poor & Good & Poor \\

\multirow{2}{*}{} ABC & Average & Good & Good & Poor & Poor \\

\multirow{2}{*}{} ALO & Average & Good & Poor & Good & Poor \\

\multirow{2}{*}{} BWO & Average & Good & Poor & Good & Poor \\

\multirow{2}{*}{} \textbf{GTBO} & \textbf{Good} & \textbf{Good} & \textbf{Good} & \textbf{Good} & \textbf{Good} \\
\hline

\hline
\end{tabular}
\end{table*}

\section{Golden Tortoise Beetle Optimizer (GTBO)}
\label{lab:fundamentals}
\subsection{Lifestyle of Golden Tortoise Beetles}
The golden tortoise beetle is one of three species of tortoise beetles that can be found in Florida, eastern North America. They can likely be found wherever all members of the morning glory family (Convolvulaceae) and also sweet potato are found. Both larvae and adults feed on foliage~\cite{olmstead1992cost, olmstead1993effectiveness, riley1870insects}. They are known as `golden bugs' and grow to around 5.0 to 7.0 mm in length. Larvae of this type of insects have moveable anal fork, e.g. peddler, which they hold over their backs as a shield to cover their bodies and deter potential enemies~\cite{olmstead1992cost}. It is very common among golden tortoise beetles to hide and then demonstrate certain patterns, a strategy to deter their predators~\cite{olmstead1993effectiveness}.

More than 30 years ago, they became the first known insect species with the ability to extremely change their colors ranging from brownish and purplish to bright orange or gold during copulation or agitation because of optical illusion~\cite{barrows1979life}. They have two states including resting and disturbed. In the resting state, these insects have their normal color of gold which arises from a chirped multilayer reflector maintained in a perfect coherent state by the presence of humidity in the porous patches within each layer. In the disturbed state, such as holding them between fingers or applying pressure on them, they will have a low-saturated red appearance resulting from the destruction of this reflector by the expulsion of the liquid from the porous patches, turning the multilayer into a translucent slab that leaves an unobstructed view of the deeper-lying, pigmented red substrate~\cite{vigneron2007switchable}.

\begin{figure}[htp!]
\centering
\includegraphics[scale=0.5]{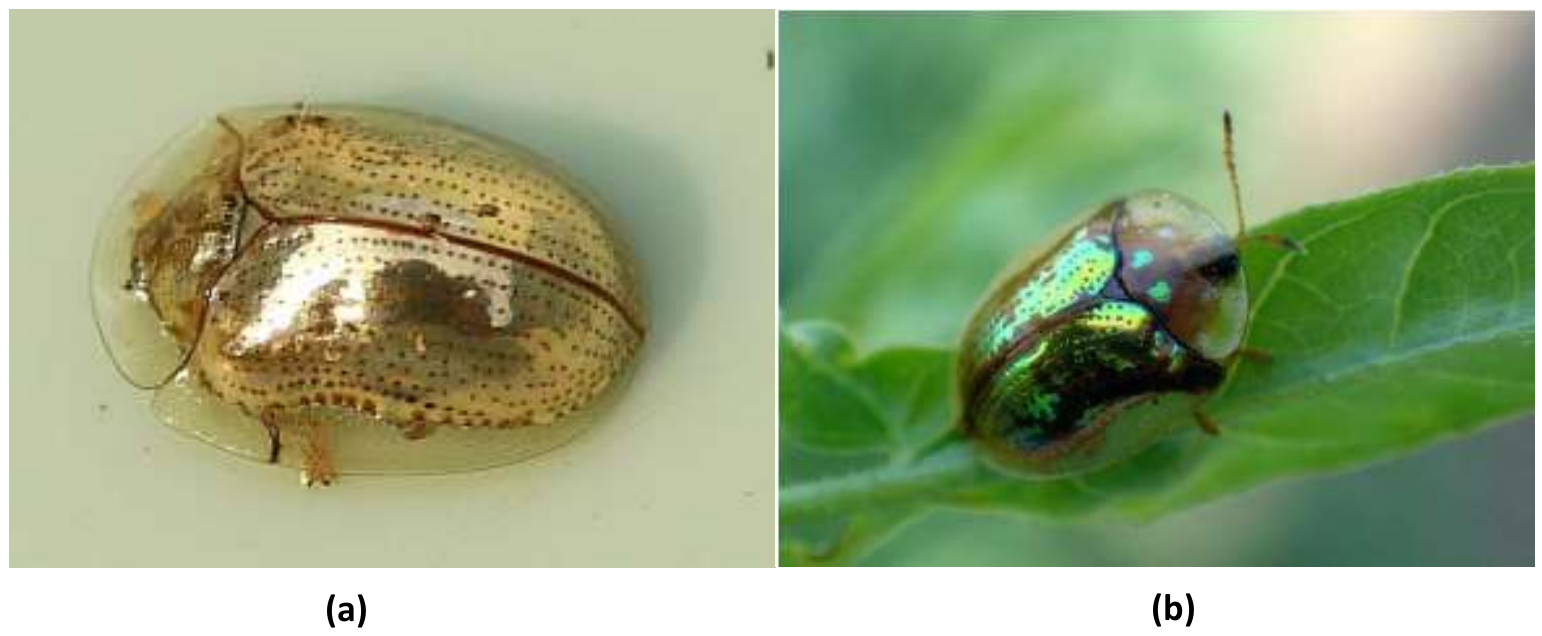}
\caption{(a) Adult golden tortoise beetle (b) adult spotted tortoise beetle.}
\label{fig:beetle}
\end{figure}

Due to the metallic quality and glare gold color of these insects, it is generally difficult for birds to see them. By changing from golden to orange and spotty, they can mimic lady beetles and with the different forms of defense, birds cannot recognize the difference~\cite{barrows1979life}. Figure~\ref{fig:beetle} shows the adult golden and spotty tortoise beetles. The time of changing color from dull red and spotty to golden is when they signal to the opposite sex to prove that they are mature, which means beetles that are not mature enough cannot produce golden color. Professor of biology Edward M. Barrows from Georgetown University, realized that these insects can not only last their copulation anywhere from 15 to 583 minutes but also alter their color in as quickly as two minutes~\cite{barrows1979life}. Another important observation is that these beetles usually have a brilliant metallic golden color when first seen on a plant, but the golden appearance is lost when they are dead and dry. They are naturally red-brown with spots when they are dead, and they return to golden or red when they are brought back to normal temperature. Elimination of moisture from the cuticle causes them to lose their metallic color and turn to red, the same reddish appearance when they are in a disturbed state. A further observation is a tense and competitive relationship among tortoise beetles, which is called `arms-race'.

The lifestyle of golden tortoise beetles has inspired us to come up with a novel optimization method that mathematically models their color-switching behavior for attracting the opposite sex to mating and reproduction and their survival mechanism for protecting themselves from predators. In this research, each beetle represents a solution or an individual. The GTBO utilizes the reproduction concept to generate solutions and the survival mechanism to determine which solutions will get to the next iteration. In the following sections, we will present the two key operators, followed by the optimization algorithm.

\subsection{Inspired Operators}
\begin{figure}[htp!]
\centering
\includegraphics[scale=0.5]{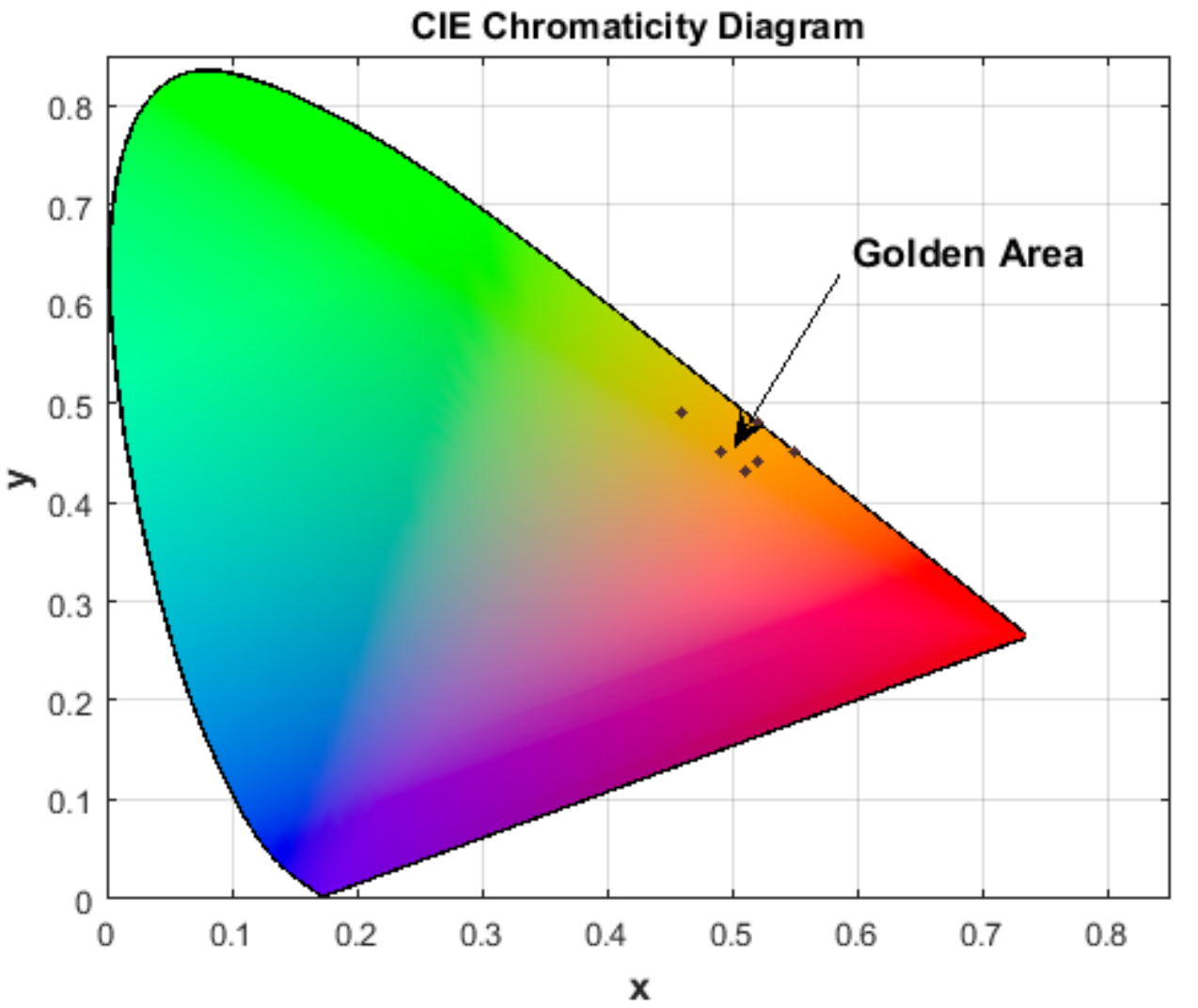}
\caption{Chromaticity coordinates.}
\label{fig:chromcor}
\end{figure}

The color switching operator is modeled after a golden tortoise beetle's behavior of changing its color when mating and at the time of disturbance. The color change is done by changing the liquid content of the interference layer under the pressure of fluid injections by the golden tortoise beetle, for example, changing the refractive index will change the dominant reflected wavelength~\cite{lenau2008colours}. To formulate the changing color operation, we use the concepts of the reflective index and the length of wavelength in thin layer interference. Consider $x$ and $y$ as two materials that are placed on top of each other like a stack. The optical thickness for each layer is $1/4$ wavelength. In other words, $d_x.n_x$=$d_y.n_y$ where $d_x$ and $d_y$ denote the layer thickness and $n_x$ and $n_y$ are the reflective indices~\cite{lenau2008colours}. The assumed reflected color can be described in Equation~\ref{eq:currbest1} when we have stacks placed in the higher reflective index.

\begin{equation}
\label{eq:currbest1}
h\cdot \lambda = (d_x\cdot n_x\cdot cos(\theta_x)+d_y\cdot n_y\cdot cos(\theta_y))\text{,}
\end{equation}

\noindent where $h$ is a constant value, $\lambda$ denotes the wavelength of the reflected light, and $\theta_x$ and $\theta_y$ are the normal angles. In order to determine the dominating wavelength ($\lambda$), we adopt Vigneron's method~\cite{vigneron2006spectral} defined for a multilayer reflector containing two alternating different layers, as formulated by Equation~\ref{eq:currbest2}.

\begin{equation}
\label{eq:currbest2}
\lambda= \frac{2\alpha \sqrt{\bar\phi^2-sin^2(\theta_z)}} {k}\text{,}
\end{equation}

\noindent where $\lambda$ denotes the wavelength, $\bar\phi$ is the mean of the reflective index, $\alpha$ indicates the thickness of the mentioned layers, $k$ is a constant integer number, and $\theta_z$ is a normal angle. Mature golden tortoise beetles utilize these mechanisms to change colors for mating and reproducing the next generations. Figure~\ref{fig:chromcor} depicts the visual aspect of gold in the chromaticity coordinates reflected in the yellow hue region of the CIE diagram. GTBO uses the color switching operator to produce new generations.

The survival operator is modeled after a beetle's protective mechanism against predators. Female beetles usually lay eggs in clusters in the places such as stems of the host leaves when they become yellowish or reddish larvae within approximately ten days. The larvae collect their shed skins using a structure called anal fork which works as a fecal shield to protect them against predators such as ants. However, some of these larvae become mature but many may not~\cite{barrows1979life, capinera2015golden}. For simplicity in generating the survived beetles, this study utilized a new crossover operator to generate $S_p$ percent of these beetles in the algorithm.

\subsection{Generating Initial Solutions}
The successfulness (efficiency) of each beetle in attracting the opposite sex via changing the color to gold and in protecting the larvae from predators has a great impact on its reproduction of the next generation. To solve an optimization problem, the problem variables need to be displayed as a matrix and in GTBO, the matrix is called a ``position". The more successful a beetle is in mating and in protecting larvae, the more efficient a solution is. Each solution converges to an optimal value over an evolutionary process as a female beetle will be attracted by a golden male beetle. The more golden the male beetle is, the more the female will change its position, and the closer it will move towards the male. In an optimization problem with $n$ dimensions, the problem variables can be represented as a $m \times n$ position matrix as follows:

\begin{equation*}
P_{beetle} =
\begin{pmatrix}
b_{1,1} & b_{1,2} & \cdots & b_{1,n} \\
b_{2,1} & b_{2,2} & \cdots & b_{2,n} \\
\vdots & \vdots & \ddots & \vdots \\
b_{m,1} & b_{m,2} & \cdots & b_{m,n}
\end{pmatrix}
\end{equation*}

\noindent where $bi,j$ is the value of the $\textit{j-th}$ variable (dimension) of the $\textit{i-th}$ beetle, $m$ is the number of beetles, and $n$ is the number of variables. The values of the defined matrix are float values.

The fitness value of each golden tortoise beetle is obtained by evaluating the cost function. Thus, the profit function of the proposed algorithm can be defined as follows:

\begin{equation*}
F_{beetle} =
\begin{pmatrix}
f ([b_{1,1} & b_{1,2} & \cdots & b_{1,n}]) \\
f([b_{2,1} & b_{2,2} & \cdots & b_{2,n})] \\
\vdots & \vdots & \ddots & \vdots \\
f([b_{m,1} & b_{m,2} & \cdots & b_{m,n})]
\end{pmatrix}
\end{equation*}

\noindent where $F_{beetle}$ shows the fitness value for each beetle and $f$ denotes the objective function. This profit function can be defined both as a minimization or maximization function.

The algorithm utilizes a random operator to initialize the first beetle. Real-valued amounts are uniformly initialized randomly between the lower and upper bounds of the dimensions $[X_{min}, X_{max}]$ where $X_{min}={{X^1}_{min},\cdots, X^D_{min}}$ and $X_{max}={X^1_{max},\cdots,X^D_{max}}$.

Equation~\ref{eq:initial} defines the initial value of the $\textit{k-th}$ parameter in the $\textit{i-th}$ beetle at the generation $G=1$ for a problem with $D$ dimensions:
\begin{equation}
\label{eq:initial}
x^k_{i,1}=x^k_{min}+rand(0,1)\cdot(x^k_{max}-x^j_{min}), k=1,2,\cdots, D.
\end{equation}

This type of initialization has also been used in different evolutionary algorithms such as differential evolution (DE) algorithm~\cite{tarkhaneh2019adaptive}.

\subsection{Switching Color Operator}
The solutions for the number of mature beetles are generated using the following equations (Equations~\ref{eq:swcolor},$\cdots$,~\ref{eq:engloss4}).
\begin{equation}
\label{eq:swcolor}
V^G_{i} = X^G_{i} +S_{color}\cdot(X_{r_1}^G-X^G_{best} ) \text{,}
\end{equation}

\noindent where $X^G_i$ is the position of current female beetle in generation $G$, which moves towards the golden male beetle $ X_{r_1}$ in generation $G$ that has a golden color with a value determined by the color switching operator $S_{color}$. In particular, $r_1$ is a randomly generated integer in $[1, NP]$ excluding $i$, where $NP$ is the number of beetle populations, while $X^G_{best}$ is the solution with the best fitness at the generation $G$.

The female beetle changes its position in order to mate with the golden beetle showing an attractive golden color and reproduce the next generation. The best-achieved solution in each generation is preserved as it is subtracted from other parts of the generated solutions. The value of $S_{color}$ is mathematically modeled using Equations~\ref{eq:engloss3} and~\ref{eq:engloss4}:
\begin{equation}
\label{eq:engloss3}
S_{color}= (d_x\cdot n_x\cdot cos(\theta_x)+d_y\cdot n_y\cdot cos(\theta_y))+(h\cdot \lambda)\text{,}
\end{equation}

\noindent where:

\begin{equation}
\label{eq:engloss4}
\resizebox{.42\hsize}{!}{$%
\begin{cases}
d_x, n_x=& Randn()\\
\\
d_y, n_y =&Randn()\cdot \beta \\
\\
\bar\phi=& Cauchy(\sigma, \mu) \\
\\
\theta_x,\theta_y=& \beta\\
\\
\alpha, k, h=& rand()\\
\\
\theta_z=& 2\cdot \pi\cdot rand()\\
\end{cases}
$%
}%
\end{equation}

In particular, $Randn()$ is a normal random function which generates numbers in $[1, n]$, $\beta$ is a uniform random function which generates numbers in the range of $[0.1, 0.9]$, $rand()$ is a random number generator between 0 and 1, and Cauchy is the standard Cauchy distribution where $\sigma$ is the location parameter and $\mu$ is a scaling parameter, which are initialized to 0.5 and 0.2 respectively. The values generated by the Cauchy distribution is initialized to 0.5 when it is outside the lower bound of zero and upper bound of 1 in the algorithm. Furthermore, variables $h$, $\lambda$, $\bar\phi$, $\alpha$, $k$, and $\theta_z$ are defined in Equations~\ref{eq:currbest1} and~\ref{eq:currbest2}.

\subsection{Survival Operator}
According to the above-mentioned details about the eggs laid by tortoise beetles, some beetles' eggs will survive due to the efficiency of their protective strategies for deterring predators. In GTBO, for the sake of simplicity yet without losing generality, a crossover operator is considered for producing $Sp$ percent of the survived beetles, which will change to the larvae in the future and then become mature beetles. The survival operator is defined using Equations~\ref{eq:Reproduce} and~\ref{eq:engloss5}.
\begin{equation}
\label{eq:Reproduce}
\resizebox{.7\hsize}{!}{$%
\begin{cases}
Beetle_1=& \alpha \cdot x_{r_1}+(1-\alpha) \cdot (x_{r_2}-\sigma_1)\\
\\
Beetle_2=& \alpha \cdot x_{r_2}+(1-\alpha) \cdot(x_{r_1}-\sigma_2)\\
\end{cases}
$%
}%
\end{equation}

\noindent where $x_{r_1}$ and $x_{r_2}$ are two randomly chosen solutions in the range of $[1, NP]$ where $NP$ is the total number of populations, and $\alpha$ is an uniform random number between [0, 1]. Variables $\sigma_1$ and $\sigma_2$ are defined in Equation~\ref{eq:engloss5}:
\begin{equation}
\label{eq:engloss5}
\resizebox{.5\hsize}{!}{$%
\begin{cases}
\sigma_1=& (1-\epsilon) \cdot (x_{best}-x_{r_1})\\
\\
\sigma_2=& (1-\epsilon) \cdot (x_{best}-x_{r_2})\\
\\
\epsilon=& \frac{\alpha \cdot \eta}{|p|^\beta}

\end{cases}
$%
}%
\end{equation}

\noindent where $x_{best}$ is the best achieved solution so far, $\eta$ and $p$ are normal numbers associated with the size of solution $x_{r_1}$, $\beta$ is a continuous uniform random number generated between [0.1, 0.5], and $\epsilon$ is a variable which is determined based on the values of $\alpha$, $\beta$, $\eta$, and $p$. Although there are some other selection operator strategies such as Roulette wheel selection method~\cite{yu2016improved}, the proposed crossover operator used a random selection strategy for choosing two beetles for generating new solutions. 

\subsection{The GTBO Algorithm}
\begin{algorithm}
\fontsize{9pt}{9pt}
\selectfont
\KwData{$X$,$ M_r (MatureRate)$, $S_r (SurvivalRate)$ }
\KwResult{$X$ \tcp{return new generated beetle population}}
$X \leftarrow \{x_1, x_2 ... x_n\}$ \tcp{initialize solutions randomly and determine fitness values using Eq.~\ref{eq:initial}}
$x^* \leftarrow best(X)$ \tcp{assign current best}
$it \leftarrow 0$ \tcp{initial iteration variable}
$NP \leftarrow $k$ $ \tcp{total number of beetle population}
$NFE \leftarrow 0$ \tcp{curent value of function evaluation}
$MP \leftarrow M_r*NP$ \tcp{number of matured beetle population}
$SP \leftarrow S_r* NP$ \tcp{number of Survival beetle population}
\Begin{

\While{Stop criterion not satisfied ($NFE<= TotalNFE$)}{

\For{$i\leftarrow 1$ \KwTo $MP$ }{
\For{each beetle $X_i$ }{
Generate solution $V_i^G$ by Eq.~\ref{eq:swcolor} \tcp {Switching Color Operator}
Evaluate the fitness value of generated solution\\
Store the generated solution in Mature Population.

\label{adjustment}
}
}
\For{$i\leftarrow 1$ \KwTo $SP$ }{
\For{each beetle $X_i$ }{
Select two solution randomly.
Generate solutions using Eq.~\ref{eq:Reproduce} \tcp {Reproduction operator}
Evaluate the fitness value of generated solution\\
Store the generated solution in survival population.

\label{adjustment}
}
}
start again from line 8 if the stopping criterion not satisfied
}
Merge the beetle population\\
Sort the population based on the fitness value and select the best NP ones.\\
Output the result.
}
\caption{GTBO}
\label{alg:GTBO}
\end{algorithm}

According to the pseudocode in Algorithm~\ref{alg:GTBO}, the GTBO starts with the initialization phase where the key control parameters such as the $M_r$ (mature rate) and $S_r$ (survival rate) are defined. The values of these control parameters should be defined in the range of (0, 1]. The number of mature beetles is determined based on the whole number of the beetle population $NP$ through the multiplication of the mature rate by $NP$. Similarly, the number of survived beetles are determined according to the survival rate and the number of beetle's population. After the initialization, the algorithm starts the main loop where it continues until the stopping criterion is satisfied.

In the main loop, first, the switching color mechanism is used to generate solutions for the number of mature beetles. Every mature beetle can change its color to golden in order to attract and mate with the opposite sex to improve its chance for reproduction. All the generated solutions are stored in the repository called the matured population. After this stage, the next operator starts and generates solutions using the number of survival beetles, which then change to matured beetles. The generated solutions at this stage are also stored in a repository called survived population. Finally, the algorithm iterates the main loop again and checks the stopping condition, and when it is satisfied it merges the population of matured beetles with the survived beetles and selects the best-generated solution and outputs the result. GTBO's main characteristics can be summarized as follows:

\begin{framed}
\noindent
1) The color switching operator and the survival operator help the algorithm guarantee balance between exploration and exploitation due to the random selection mechanism. \\
\noindent
2) Due to the random selection of some solutions in the population-based algorithm, there is a high probability of local optima avoidance. \\
\noindent
3) The algorithm has a good population diversity as matured and survived beetle population is merged and then sorted. \\
\noindent
4) In every iteration, the best solution will preserve, update, and compare with generated solutions (elite). \\
\noindent
5) The algorithm has few parameters to adjust.
\end{framed}

\section{Experiments}
\label{lab:experiments}

\subsection{Benchmark functions and baseline algorithms}

\begin{figure*}[htp!]
\centering
\includegraphics[scale=0.8]{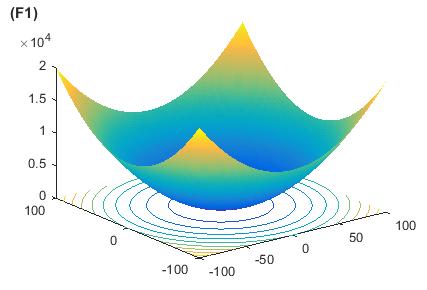}
\includegraphics[scale=0.8]{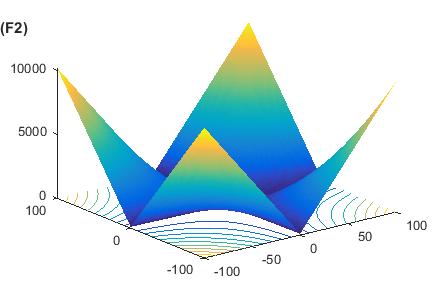}
\includegraphics[scale=0.8]{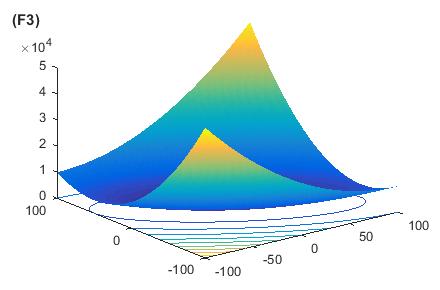}
\includegraphics[scale=0.8]{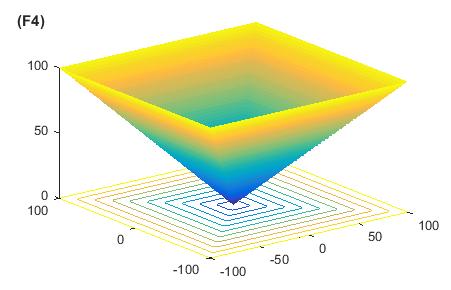}
\includegraphics[scale=0.8]{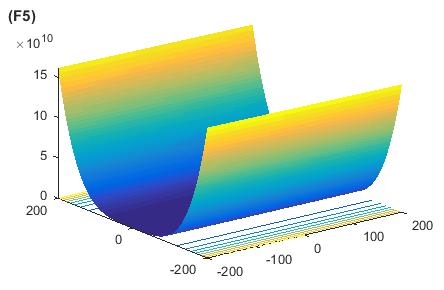}
\includegraphics[scale=0.8]{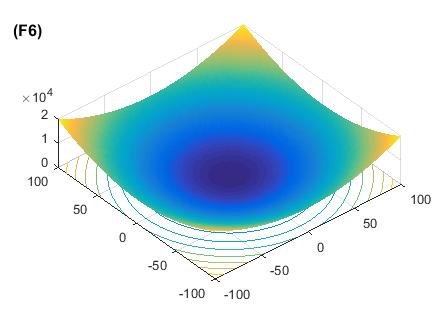}
\includegraphics[scale=0.8]{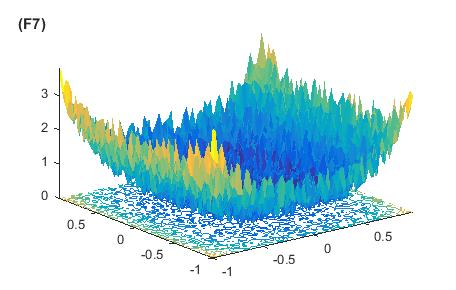}

\caption{F1-F7 represnts the 3-D versions of unimodal functions. }
\label{fig:unifunc}
\end{figure*}

\begin{figure*}[htp!]
\centering
\includegraphics[scale=0.8]{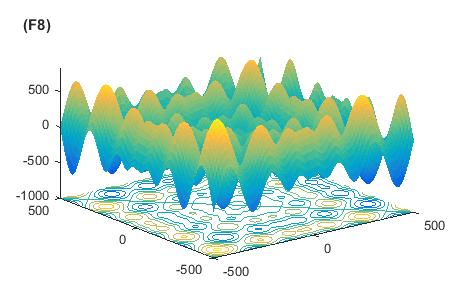}
\includegraphics[scale=0.8]{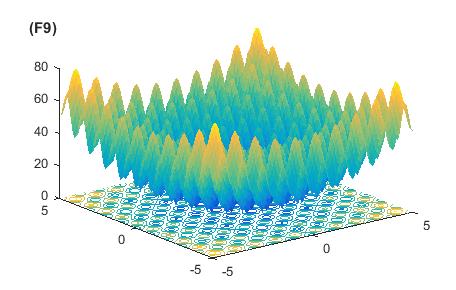}
\includegraphics[scale=0.8]{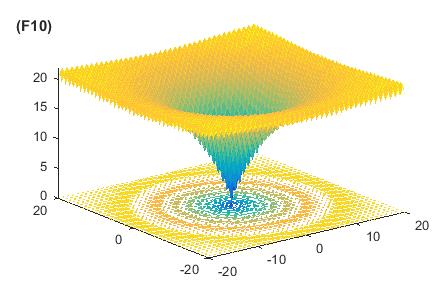}
\includegraphics[scale=0.8]{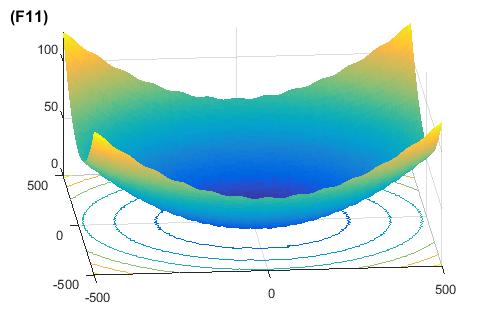}
\includegraphics[scale=0.8]{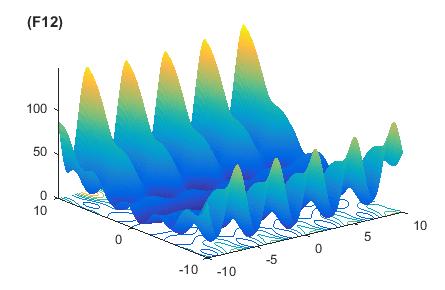}
\includegraphics[scale=0.8]{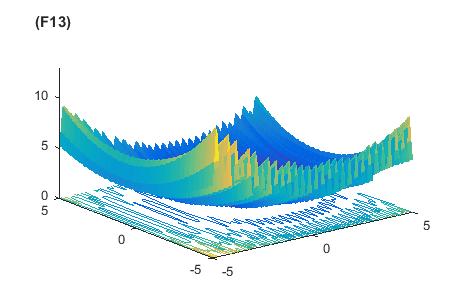}

\caption{F8-F13 represnts the 3-D versions of multimodal functions. }
\label{fig:multifunc}
\end{figure*}

In this section, several benchmark functions are employed to evaluate the performance of the proposed algorithm by using different criteria. In this regard, three classes of benchmark test functions are employed to prove the efficiency of the proposed algorithm~\cite{digalakis2001benchmarking, molga2005test, liang2005novel}. The first class is unimodal functions described in Table~\ref{tab:uni}. In this class, there is one single optimum where the efficiency of the algorithm can be assessed through the exploitation and the convergence speed. Another class is to do with multimodal functions where it consists of more than one optimum. This class is more complicated compared to the unimodal with which the proposed algorithm can be evaluated in terms of both exploration and avoidance of local minima~\cite{mirjalili2015ant}. This type of class is listed in Table~\ref{tab:multi}~\cite{digalakis2001benchmarking, molga2005test, liang2005novel}. The third class of test functions is hybrid composite functions described in CEC 2005 special session~\cite{suganthan2005problem}. One algorithm should properly make a trade-off between exploration and exploitation to achieve efficient results in composite functions~\cite{mirjalili2015ant}. Thus, this class of functions, which are listed in Table~\ref{tab:composfunc}, tests the ability of the proposed algorithm in terms of the balance between both exploration and exploitation. The 3-dimensional shapes of the unimodal and multimodal functions are presented in Figures~\ref{fig:unifunc} and \ref{fig:multifunc}. Regarding the baseline algorithms, this study employed standard GA~\cite{holland1992genetic}, ABC~\cite{karaboga2009comparative}, PSO~\cite{poli2007particle}, Black Widow Optimization (BWO) algorithm~\cite{hayyolalam2020black}, and Ant Lion Optimizer (ALO) algorithm~\cite{mirjalili2015ant}.

\subsection {Evaluation criteria and parameter settings}
The number of population ($NP$) for all the algorithms is initialized to 100. As a fair stopping condition, this research utilizes the total number of function evaluations (TotalNFE) initialized to 100,000 for all the algorithms. All algorithms performed 30 independent runs using Matlab 2016 on a computer with Intel Core i3, 2.5 GHz, 4 GB RAM running Windows 7. Table~\ref{tab:parameters} shows the parameter settings in detail. The parameter settings of the ALO algorithm are determined based on its original paper. Some criteria such as the best, mean fitness value, and standard deviation (std) are employed to show the results of all algorithms. However, in order to show the superiority of the proposed algorithm, this research conducted the Wilcoxon signed rank-sum test to show that the proposed algorithm is statistically significant with respect to the baseline algorithms~\cite{wilcoxon1945individual}. The details of this test will be discussed in the later sections.

\section{Results and Discussions}
\label{lab:results}

\subsection{Results on unimodal functions}
Tables~\ref{tab:uni1},~\ref{tab:wilcoxonuni1}, and~\ref{tab:uni2} reveal the results related to the unimodal test functions. In these tables, best denotes the best-achieved fitness value over 30 independent runs, mean shows the average value of fitness value, and STD indicates the standard deviation of obtained results. The superior results of the GTBO are shown in boldface.

\begin{table*}
\fontsize{8pt}{12pt}
\selectfont
\centering
\caption{ Unimodal benchmark functions. (Dim:
Dimension, range: Lower and upper bound of problem, $f_{min}$: global minimum value of function
).}
\label{tab:uni}
\begin{tabular}{llllllll}
\hline

Function \ \ & Dim & Range & $f_{min}$ \\
\hline

$F_1(x)=\sum_{i=1}^{n} x_i^{2}$ & 30, 200 & [-100, 100] & 0\\

$F_2(x)=\sum_{i=1}^{n} |x_i|+\displaystyle\prod_{i=1}^{n} |x_i|$ & 30, 200 & [-10, 10] & 0\\

$F_3(x)=\sum_{i=1}^{n} (\sum_{j-1}^{i} x_j)^2$ & 30, 200 & [-100, 100] & 0\\

$F_4(x)=\displaystyle max_{i}\{|x_i|, 1\leq i \leq n\}$ & 30, 200 & [-100, 100] & 0\\
$F_5(x)=\sum_{i=1}^{n-1} [100(x_{i+1}-x_i^2)^2+(x_i-1)^2]$ & 30, 200 & [-30, 30] & 0\\
$F_6(x)=\sum_{i=1}^{n} ([x_i+0.5])^2$ & 30, 200 & [-100, 100] & 0\\
$F_7(x)=\sum_{i=1}^{n} ix_i^4+random[0,1)$ & 30, 200 & [-1.28, 1.28] & 0\\
\hline

\end{tabular}
\end{table*}

\begin{table*}
\fontsize{6pt}{14pt}
\selectfont
\centering
\caption{ Multimodal benchmark functions. (Dim:
Dimension, range: Lower and upper bound of problem, $f_{min}$: global minimum value of function).}
\label{tab:multi}
\begin{tabular}{llllllll}
\hline

Function \ \ & Dim & Range & $f_{min}$ \\
\hline

$F_8(x)=\sum_{i=1}^{n} -x_i sin(\sqrt{|x_i|})$ & 30, 200 & [-500, 500] & $-418.9829\times Dim$\\

$F_9(x)=\sum_{i=1}^{n} [x_i^2-10 cos(2\pi x_i)+10]$ & 30, 200 & [-5.12, 5.12] & 0\\

$F_{10}(x)=-20 exp (-0.2\sqrt{\frac{1}{n}\sum_{i=1}^{n}x_i^2})-exp(\frac{1}{n}\sum_{i=1}^{n}cos(2\pi x_i))+20+e$ & 30, 200 & [-32, 32] & 0\\

$F_{11}(x)=\frac{1}{4000}\sum_{i=1}^{n}x_i^2-\prod_{i=1}^{n}cos(\frac{x_i}{\sqrt{i}})+1$ & 30, 200 & [-600, 600] & 0\\
$F_{12}(x)=\frac{\pi}{n}\{10sin(\pi y_1)+\sum_{i=1}^{n-1}(y_i-1)^2 [1+10sin^2(\pi y_{i+1})]+(y_n-1)^2 \}+\sum_{i=1}^{n} u(x_i,10,100,4)$ & 30, 200 & [-50, 50] & 0\\
$F_{13}(x)=0.1\{sin^2 (3\pi x_1)+\sum_{i=1}^{n} (x_i-1)^2 [1+sin^2(3\pi x_i+1)]+(x_n-1)^2 [1+sin^2(2\pi x_n)] \}
+ \sum_{i=1}^{n} u(x_i,5,100)$ & 30, 200 & [-50, 50] & 0\\

\hline

\end{tabular}
\end{table*}

\begin{table*}
\fontsize{8pt}{12pt}
\selectfont
\centering
\caption{ Hybrid composition benchmark functions (CEC2005) used for experimental study (Dim:
Dimension, Min, Max: Limits of search space, H: Hybrid, C: Composition
).}
\label{tab:composfunc}
\begin{tabular}{llllllll}
\hline

Function No \ \ & Name & Type & Min & Max & Dim \\
\hline

F15 & hybrid comp. Fn 1&HC& -5& 5& 30\\

F16 &Rotated hybrid comp. Fn 1&HC& -5& 5& 30\\

F17 & Rotated hybrid comp. Fn 1 with noise & HC&-5& 5& 30\\

F18 &Rotated hybrid comp. Fn 2& HC& -5& 5&30\\

F19 &Rotated hybrid comp. Fn 2 with narrow global optimal& HC&-5& 5&30\\

F20 & Rotated hybrid comp. Fn 2 with the global optimum& HC &-5& 5& 30\\

F21 &Rotated hybrid comp. Fn 3& HC&-5& 5& 30\\

F22 &Rotated hybrid comp. Fn 3 with high condition number matrix& HC&-5& 5& 30\\

F23 &Non-continuous rotated hybrid comp. Fn 3& HC&-5& 5& 30\\

F24 &Rotated hybrid comp. Fn 4& HC&-5& 5& 30\\

F25 &Rotated hybrid comp. Fn 4 without Bounds& HC&-2& 5& 30\\

\hline

\end{tabular}
\end{table*}

\begin{table*}
\begin{center}
\fontsize{8pt}{12pt}
\selectfont
\caption{Parameter settings of the employed algorithms}
\resizebox{.5\hsize}{!}{$%
{\begin{tabular}{@{}ll@{}ll@{}}
\hline
Algorithm &Parameter & Description \\
\hline
GA & $ pc=0.67 $ & Crossover Rate\\
& $ Mc=0.33 $ & Mutation Rate\\
PSO & $ w=2 $ & Inertia Weight\\
& $ bg=2.2$ & Best global experience\\
& $ wdamp = 0.98$ & \\
& $bp=2.4$ & Best personal experience \\
BWO & $ PP=0.6 $ & Procreate Rate\\
& $CR=0.44$ & Cannibalism Rate\\
ABC & $No. \ \ of \ \ food \ \ sources=NP/2$ & \\
& $Limit=15$ & \\
GTBOA & $M_r \ \ (Mature \ \ Rate)=0.4$ & Rate. of Matured Beetles \\
& $S_r \ \ (Survival \ \ Rate)=0.2$ & Rate. of Survived Beetles\\
$TotalNFE = 100, 000$ & No. of Function Evaluations\\

\hline
\label{tab:parameters}
\end{tabular}}
$%
}%
\end{center}
\end{table*}

\begin{table*}
\fontsize{8pt}{12pt}
\selectfont
\centering
\caption{ Unimodal benchmark functions results of all employed algorithms for 30 dimensions.}
\label{tab:uni1}
\begin{tabular}{llllllll}
\hline

Function No \ \ & Criteria & GA & PSO & ABC & ALO & BWO & GTBOA \\
\hline

F1 & Best & 1.2131e-06 & 108.9183 & 3.0858 & 1.7124 & 0.0011 & \textbf{2.2912e-19}\\
& Mean & 1.6048 & 375.9294 & 14.3653 & 1.8804 & 0.0029 & \textbf{3.4636e-18}\\
& STD & 3.1829 & 240.5136 & 10.8668 & 0.2171 & 0.0019 & \textbf{4.2131e-18}\\

F2 & Best & 7.5194e-05 & 2.1456 & 1.5127 & 1.1277 & 0.0031 & \textbf{5.9375e-12}\\
& Mean & 0.1006 & 10.6079 & 2.4668 & 41.7537 & 0.0072 & \textbf{1.7371e-11}\\
& STD & 0.1373 & 7.5088 & 0.7499 & 51.9005 & 0.0066 & \textbf{1.2793e-11}\\

F3 & Best & 445.2838 & 1.4940e+04 & 2.3119e+04 & 36.6311 & 141.9181 & \textbf{4.1325}\\
& Mean & 689.4357 & 2.4175e+04 & 2.8024e+04 & 44.9914 & 230.9100 & \textbf{14.7848}\\
& STD & 223.5618 & 4.1746e+03 & 2.3374e+03 & 7.2157 & 84.8917 & 12.5700\\

F4 & Best & 4.6389e-04 & 32.0965 & 56.8363 & 0.9720 & 0.6664 & 0.3167\\
& Mean & 0.5985 & 36.4333 & 68.2251 & 2.7303 & 1.3689 & 1.3985\\
& STD & 0.8674 & 3.1428 & 4.5274 & 1.6668 & 0.4055 & 1.0517\\

F5 & Best & 162.9445 & 4.1259e+04 & 564.6546 & 37.5728 & 100.2362 & \textbf{24.3109}\\
& Mean & 1.3213e+03 & 9.8076e+04 & 905.6610 & 303.8040 & 216.1671 & \textbf{45.2228}\\
& STD & 1.2222e+03 & 7.2228e+04 & 478.0250 & 499.1320 & 76.1047 & \textbf{28.4930}\\

F6 & Best & 3.3299e-04 & 66.9998 & 2.1334 & 1.6228 & 4.3832e-04 & \textbf{7.7615e-20}\\
& Mean & 0.6190 & 279.7871 & 9.5629 & 1.9982 & 0.0019 & \textbf{3.3433e-18}\\
& STD & 0.8442 & 164.8274 & 9.4036 & 0.2518 & 0.0013 & \textbf{5.3386e-18}\\

F7 & Best & 0.0027 & 0.2594 & 0.5100 & 0.0057 & 0.0014 & 0.0054\\
& Mean & 0.0068 & 0.9902 & 1.2994 & 0.0093 & 0.0040 & 0.0132\\
& STD & 0.0027 & 1.0903 & 0.5285 & 0.0038 & 0.0020 & 0.0060\\
\hline

\end{tabular}
\end{table*}

\begin{table*}
\fontsize{8pt}{12pt}
\selectfont

\centering
\caption{ Statistical analysis (Wilcoxon rank-sum test) on unimodal test functions with 30 dimensions}
\label{tab:wilcoxonuni1}
\resizebox{0.8\hsize}{!}{$%
\begin{tabular}{ll@{\qquad}llll@{\qquad}llll@{\qquad}llllllll@{\qquad}}

\toprule {Function No.}
\multirow{16}{*}{\raisebox{-\heavyrulewidth}} & \multicolumn{4}{c}{ GTBOA vs GA} & \multicolumn{4}{c} \ \ \ { GTBOA vs PSO} & \multicolumn{4}{c} \ \ \ {GTBOA vs ABC } & \multicolumn{4}{c} \ \ \ { } \\
\cmidrule{3-5}
\cmidrule{7-9}
\cmidrule{10-13}
\centering
& & pval & T+ & T-& winner & pval & T+ & T-& winner& pval & T+ & T-& winner \\
\midrule
F1 & & 0.0118$(<0.05^+)$ & 751 & 239 & + & 1.0202e-187$(<0.05^+)$ & 957 & 33 & +& 1.0029e-123$(<0.05^+)$ & 693 & 297 & + \\
F2 & & 0.3843$(>0.05^+)$ & 565 & 425 & - & 4.4929e-161$(<0.05^+)$ & 971 & 19 & + & 9.4246e-152$(<0.05^+)$ & 852 & 138 & +\\
F3 & & 5.5314e-05$(<0.05^+)$ &535 & 455 & + & 6.9057e-236$(<0.05^+)$ & 936 & 54 & +& 1.6768e-283$(<0.05^+)$ & 982 & 8 & + \\
F4 & & 0.4210$(>0.05^+)$ & 0 & 990 & - & 1.0075e-240$(<0.05^+)$ & 932 & 58 & +& 3.9268e-320$(<0.05^+)$ & 990 & 0 & + \\
F5 & & 1.6447e-78$(<0.05^+)$ & 723 & 267 & + & 1.7986e-210$(<0.05^+)$ & 980 & 10 & +& 1.3470e-133$(<0.05^+)$ & 872 & 118 & + \\
F6 & & 2.3143e-31$(<0.05^+)$ & 611 & 379 & + & 4.1734e-186$(<0.05^+)$ & 972 & 18 & +& 2.7740e-135$(<0.05^+)$ & 884 & 106 & + \\
F7 & & 1.0620e-170$(<0.05^+)$ & 710 & 280 & + & 9.8038e-235$(<0.05^+)$ & 979 & 11 & +& 1.9982e-261$(<0.05^+)$ & 990 & 0 & + \\

\\
\hline
\\
\multirow{16}{*}{\raisebox{-\heavyrulewidth}} & \multicolumn{4}{c}{ GTBOA vs ALO} & \multicolumn{4}{c} \ \ \ { GTBOA vs BWO} & \multicolumn{4}{c} \ \ \ { } & \multicolumn{4}{c} \ \ \ { } \\
\cmidrule{3-5}
\cmidrule{7-9}

\centering
& & pval & T+ & T-& winner & pval & T+ & T-& winner \\
\\
F1 & & 1.2234e-140$(<0.05^+)$ & 843 & 147 & + & 8.5090e-19$(<0.05^+)$ & 589 & 401 & + \\
F2 & & 1.5410e-181$(<0.05^+)$ & 971 & 19 & + & 5.0797e-17$(<0.05^+)$ & 582 & 408 & + \\
F3 & & 3.3480e-06$(<0.05^+)$ & 691 & 299 & + & 2.0306e-11$(<0.05^+)$ & 610 & 380 & + \\
F4 & & 3.4211e-46$(<0.05^+)$ & 900 & 90 & + & 5.9719e-27$(<0.05^+)$ & 655 & 335 & + \\
F5 & & 7.2719e-101$(<0.05^+)$ & 939 & 51 & + & 4.7413e-69$(<0.05^+)$ & 706 & 284 & + \\
F6 & & 1.0515e-154$(<0.05^+)$ & 936 & 54 & + & 3.5595e-26$(<0.05^+)$ & 611 & 379 & + \\
F7 & & 0.0820$(>0.05^+)$ & 345 & 645 & - & 0.3214$(<0.05^+)$ & 102 & 888 & + \\
\\

\bottomrule
\end{tabular}
$%
}%
\end{table*}

\begin{table*}
\fontsize{8pt}{12pt}
\selectfont
\centering
\caption{ Unimodal benchmark functions results of all employed algorithms for 50 dimensions.}
\label{tab:uni2}
\begin{tabular}{llllllll}
\hline

Function No \ \ & Criteria & GA & PSO & ABC & ALO & BWO & GTBOA \\
\hline

F1 & Best & 1.0000e-03 & 2.1415e+03 & 104.5097 & 6.1849 & 3.1374 & \textbf{6.8041e-09}\\
& Mean & 0.2401 & 3.9719e+03 & 262.8478 & 7.5865 & 9.6358 & \textbf{9.1314e-08}\\
& STD & 0.3944 & 1.0605e+03 & 185.3239 & 0.9823 & 4.0663 & \textbf{8.2333e-08}\\

F2 & Best & 4.0000e-04 & 30.8469 & 4.1226 & 3.2234 & 0.5971 & \textbf{1.6904e-06}\\
& Mean & 0.3842 & 79.0294 & 5.8825 & 82.7113 & 1.0424 & \textbf{5.3055e-06}\\
& STD & 0.6819 & 33.8814 & 1.0043 & 95.6385 & 0.3665 & \textbf{2.9952e-06}\\

F3 & Best & 1.2141e+03 & 6.6226e+04 & 6.7060e+04 & 886.5140 & 567.1294 & \textbf{1.2137e+03}\\
& Mean & 1.9825e+03 & 8.1653e+04 & 8.2528e+04 & 1.0102e+03 & 870.9053 & 3.0783e+03\\
& STD & 669.1000 & 1.0852e+04 & 8.4185e+03 & 91.0529 & 231.4373 & 1.2043e+03\\

F4 & Best & 0.0397 & 51.3212 & 77.3875 & 8.2060 & 5.6129 & 16.0928\\
& Mean & 1.8005 & 56.5572 & 80.5567 & 12.1953 & 8.4819 & 21.4955\\
& STD & 3.1948 & 2.8711 & 1.9138 & 4.1685 & 2.4756 & 3.5943\\

F5 & Best & 270.7661 & 1.2681e+06 & 2.6798e+03 & 100.1233 & 563.2197 & \textbf{73.7331}\\
& Mean & 9.6687e+03 & 3.0598e+06 & 6.1883e+03 & 191.6021 & 1.9829e+03 & \textbf{149.3488}\\
& STD & 1.2861e+04 & 1.8600e+06 & 2.7620e+03 & 98.9993 & 1.4415e+03 & \textbf{81.9379}\\

F6 & Best & 1.0000e-04 & 1.9402e+03 & 40.8407 & 5.5331 & 3.9743 & \textbf{5.6347e-09}\\
& Mean & 1.6303 & 4.1254e+03 & 210.3022 & 7.9905 & 7.7142 & \textbf{5.4944e-08}\\
& STD & 2.6221 & 1.5306e+03 & 106.4296 & 1.7843 & 2.9658 & \textbf{6.0208e-08}\\

F7 & Best & 0.0153 & 1.8539 & 4.1916 & 0.0185 & 0.0120 & 0.0283\\
& Mean & 0.0364 & 7.4114 & 7.4776 & 0.0297 & 0.0291 & 0.0556\\
& STD & 0.0236 & 9.6894 & 2.5425 & 0.0089 & 0.0134 & 0.0259\\
\hline

\end{tabular}
\end{table*}

\begin{table*}
\fontsize{8pt}{12pt}
\selectfont
\centering
\caption{ Multimodal benchmark functions results of all employed algorithms for 30 dimensions.}
\label{tab:multi1}
\begin{tabular}{llllllll}
\hline

Function No \ \ & Criteria & GA & PSO & ABC & ALO & BWO & GTBOA \\
\hline

F8 & Best & -4.1661e+03 & -3833.3258 & -4.0026e+03 & -5.7714e+03 & -4.1663e+03 & \textbf{-1.0570e+04}\\
& Mean & -3.9527e+03 & -3.3568e+03 & -3.8028e+03 & -5.5843e+03 & -3.9670e+03 & \textbf{-9.3533e+03}\\
& STD & 352.6358 & 269.1801 & 121.9016 & 134.6784 & 151.6641 & 773.4089\\

F9 & Best & 2.5011e-12 & 66.7150 & 59.1072 & 32.9624 & 0.0094 & 22.8840\\
& Mean & 0.0267 & 173.4088 & 66.0430 & 51.4169 & 1.6281 & 35.4868\\
& STD & 0.0552 & 49.5622 & 5.7181 & 18.9709 & 2.2689 & 12.0436\\

F10 & Best & 4.8116e-04 & 4.0877 & 4.9133 & 0.7608 & 0.0055 & \textbf{1.1199e-10}\\
& Mean & 0.0869 & 6.1922 & 7.3804 & 1.8218 & 0.0125 & \textbf{4.2979e-10}\\
& STD & 0.1442 & 1.2298 & 1.4983 & 0.9408 & 0.0089 & \textbf{2.4641e-10}\\

F11 & Best & 1.9055e-11 & 1.5373 & 1.0836 & 0.9664 & 0.0036 & \textbf{0.0000}\\
& Mean & 0.4100 & 5.2037 & 1.2179 & 1.0016 & 0.0586 & \textbf{0.0031}\\
& STD & 0.4439 & 2.1055 & 0.1136 & 0.0202 & 0.0688 & \textbf{0.0047}\\

F12 & Best & 5.8804e-06 & 20.3593 & 0.0348 & 3.1373 & 0.0012 & \textbf{4.4676e-09}\\
& Mean & 7.6819e-04 & 1.0549e+04 & 0.2412 & 5.7674 & 0.0411 & \textbf{2.0542e-04}\\
& STD & 0.0018 & 2.5201e+04 & 0.1640 & 2.7118 & 0.0588 & \textbf{4.9198e-04}\\

F13 & Best & 4.9435e-05 & 226.5227 & 0.3491 & 0.4885 & 0.0573 & 4.3196e-08\\
& Mean & 0.4128 & 3.2245e+05 & 0.8159 & 0.6337 & 0.3013 & \textbf{0.0015}
\\
& STD & 0.6804 & 5.7834e+05 & 0.4147 & 0.1448 & 0.2701 & \textbf{0.0038}\\
\hline

\end{tabular}
\end{table*}

\begin{table*}
\fontsize{8pt}{12pt}
\selectfont

\centering
\caption{ Statistical analysis (Wilcoxon rank-sum test) on multimodal test functions with 30 dimensions}
\label{tab:wilcoxonmulti}
\resizebox{0.8\hsize}{!}{$%
\begin{tabular}{ll@{\qquad}llll@{\qquad}llll@{\qquad}llllllll@{\qquad}}

\toprule {Function No.}
\multirow{16}{*}{\raisebox{-\heavyrulewidth}} & \multicolumn{4}{c}{ GTBOA vs GA} & \multicolumn{4}{c} \ \ \ { GTBOA vs PSO} & \multicolumn{4}{c} \ \ \ {GTBOA vs ABC } & \multicolumn{4}{c} \ \ \ { } \\
\cmidrule{3-5}
\cmidrule{7-9}
\cmidrule{10-13}
\centering
& & pval & T+ & T-& winner & pval & T+ & T-& winner& pval & T+ & T-& winner \\
\midrule
F8 & & 1.3227e-263$(<0.05^+)$ & 990 & 0 & + & 2.6385e-309$(<0.05^+)$ & 990 & 0 & +& 5.9288e-323$(<0.05^+)$ & 990 & 0 & + \\
F9 & & 0.2412$(>0.05^+)$ & 0 & 990 & - & 1.1810e-97$(<0.05^+)$ & 932 & 27 & + & 1.2306e-75$(<0.05^+)$ & 766 & 224 & +\\
F10 & & 5.3312e-09$(<0.05^+)$ & 513 & 477 & + & 3.8201e-193$(<0.05^+)$ & 972 & 18 & +& 3.1481e-214$(<0.05^+)$ & 986 & 4 & + \\
F11 & & 4.0220e-56$(<0.05^+)$ & 679 & 311 & + & 6.2604e-192$(<0.05^+)$ & 969 & 21 & +& 5.4183e-163$(<0.05^+)$ & 896 & 94 & + \\
F12 & & 0.0925$(>0.05^+)$ & 115 & 384 & - & 9.0311e-95$(<0.05^+)$ & 482 & 17 & +& 1.1016e-18$(<0.05^+)$ & 369 & 130 & + \\
F13 & & 1.0087e-26$(<0.05^+)$ & 401 & 98 & + & 5.9091e-75$(<0.05^+)$ & 473 & 26 & +& 3.3476e-10$(<0.05^+)$ & 342 & 157 & + \\

\\
\hline
\\
\multirow{16}{*}{\raisebox{-\heavyrulewidth}} & \multicolumn{4}{c}{ GTBOA vs ALO} & \multicolumn{4}{c} \ \ \ { GTBOA vs BWO} & \multicolumn{4}{c} \ \ \ { } & \multicolumn{4}{c} \ \ \ { } \\
\cmidrule{3-5}
\cmidrule{7-9}

\centering
& & pval & T+ & T-& winner & pval & T+ & T-& winner \\
\\
F8 & & 3.9598e-190$(<0.05^+)$ & 873 & 113 & + & 9.8074e-247$(<0.05^+)$ & 950 & 40 & + \\
F9 & & 2.0175e-56$(<0.05^+)$ & 884 & 106 & + & 0.0725$(>0.05^+)$ & 0 & 990 & - \\
F10 & & 5.4181e-159$(<0.05^+)$ & 936 & 54 & + & 5.3513e-25$(<0.05^+)$ & 607 & 383 & + \\
F11 & & 1.3326e-160$(<0.05^+)$ & 932 & 58 & + & 8.3918e-36$(<0.05^+)$ & 635 & 355 & + \\
F12 & & 9.4212e-65$(<0.05^+)$ & 434 & 65 & + & 0.0052$(<0.05^+)$ & 272 & 227 & +\\
F13 & & 2.5806e-33$(<0.05^+)$ & 439 & 60 & + & 0.0322$(<0.05^+)$ & 276 & 223 & + \\

\\

\bottomrule
\end{tabular}
$%
}%
\end{table*}

\begin{table*}
\fontsize{8pt}{12pt}
\selectfont
\centering
\caption{ Multimodal benchmark functions results of all employed algorithms for 50 dimensions.}
\label{tab:multi2}
\begin{tabular}{llllllll}
\hline

Function No \ \ & Criteria & GA & PSO & ABC & ALO & BWO & GTBOA \\
\hline

F8 & Best & -1.8123e+04 & -1.4103e+04 & -1.4177e+04 & -9.2607e+03 & -1.2220e+04 & -1.6429e+04\\
& Mean & -1.5776e+04 & -1.2733e+04 & -1.3103e+04 & -9.1438e+03 & -1.0800e+04 & -1.5095e+04\\
& STD & 1.7020e+03 & 804.3379 & 572.4003 & 113.0223 & 1.0467e+03 & \textbf{514.2831}\\

F9 & Best & 8.0000e-04 & 291.5818 & 142.8249 & 96.1233 & 9.4104 & 41.7882\\
& Mean & 1.6294 & 407.2185 & 169.8958 & 108.0825 & 21.9785 & 71.1063\\
& STD & 1.7271 & 82.0520 & 22.1329 & 15.0202 & 7.6228 & 21.7004\\

F10 & Best & 8.0000e-04 & 9.6225 & 7.5975 & 2.4662 & 0.5517 & \textbf{2.3944e-05}\\
& Mean & 0.0839 & 11.2567 & 10.6152 & 3.0566 & 1.0042 & \textbf{0.0016}\\
& STD & 0.0701 & 0.7568 & 1.8320 & 0.5687 & 0.3508 & \textbf{5.6849e-04}\\

F11 & Best & 6.0000e-04 & 20.7541 & 1.7991 & 1.0630 & 1.0212 & \textbf{6.8045e-09}\\
& Mean & 0.3076 & 38.2575 & 3.4333 & 1.0789 & 1.1043 & \textbf{0.0099}\\
& STD & 0.4770 & 12.6355 & 0.8731 & 0.0119 & 0.0734 & 0.0254\\

F12 & Best & 1.6813 & 3.3053e+04 & 0.4340 & 14.5439 & 1.4612 & \textbf{0.1550}\\
& Mean & 4.6357 & 2.2985e+06 & 1.0839 & 19.8078 & 4.9364 & \textbf{2.9127}\\
& STD & 1.9286 & 3.0329e+06 & 0.5251 & 6.2547 & 0.2615 & 1.9893\\

F13 & Best & 1.0000e-04 & 1.5536e+06 & 1.4779 & 4.0195 & 6.4525 & 0.9129\\
& Mean & 11.1122 & 6.9882e+06 & 5.5231 & 90.7775 & 14.1978 & \textbf{9.9584}\\
& STD & 0.1380 & 8.2902e+06 & 5.3147 & 51.2084 & 7.5786 & 6.5993\\
\hline

\end{tabular}
\end{table*}

\begin{table*}
\fontsize{8pt}{12pt}
\selectfont
\centering
\caption{ Hybrid composite benchmark functions results of all employed algorithms for 30 dimensions.}
\label{tab:compos1}
\begin{tabular}{llllllll}
\hline

Function No \ \ & Criteria & GA & PSO & ABC & ALO & BWO & GTBOA \\
\hline

F15 & Best & 261.2875 & 413.9231 & 310.3578 & 502.5654 & 520.1649 & \textbf{200.0063}\\
& Mean & 587.2951 & 586.7310 & 481.2109 & 560.0458 & 580.2370 & \textbf{404.3135}\\
& STD & 138.5073 & 71.9346 & 68.0997 & 53.8564 & 36.3407 & 184.1164 \\

F16 & Best & 294.4417 & 304.6725 & 397.8783 & 118.7518 & 291.1114 & \textbf{54.3660}\\
& Mean & 503.2913 & 424.3700 & 443.5813 & 266.0464 & 394.3834 & \textbf{110.9351}\\
& STD & 124.2395 & 102.7154 & 19.7516 & 175.5074 & 101.6714 & 37.3781\\

F17 & Best & 289.1165 & 292.1645 & 636.3725 & 230.6587 & 330.7373 & \textbf{73.2936}\\
& Mean & 487.4050 & 405.3080 & 766.5341 & 320.7117 & 500.0729 & \textbf{148.9198}\\
& STD & 101.2404 & 110.4219 & 61.6579 & 104.6620 & 147.9874 & \textbf{108.5539}\\

F18 & Best & 1.1030e+03 & 916.7264 & 955.7344 & 916.7859 & 1.0990e+03 & \textbf{885.7073}\\
& Mean & 1.1517e+03 & 959.8715 & 994.9993 & 925.5520 & 1.1258e+03 & \textbf{899.2759}\\
& STD & 29.2261 & 35.5771 & 18.7432 & 8.1187 & 15.6452 & \textbf{6.2678}\\

F19 & Best & 1.1085e+03 & 933.1690 & 954.5782 & 931.8614 & 1.0861e+03 & \textbf{898.0658}\\
& Mean & 1.1527e+03 & 984.4148 & 989.0768 & 974.8522 & 1.1234e+03 & \textbf{902.7453}\\
& STD & 25.9372 & 41.1798 & 22.6374 & 47.3148 & 15.6621 & \textbf{3.3756}\\

F20 & Best & 1.0677e+03 & 923.3271 & 967.4342 & 800.1327 & 1.0976e+03 & 890.1426\\
& Mean & 1.1463e+03 & 956.9104 & 997.2200 & 877.0547 & 1.1253e+03 & 900.2646\\
& STD & 38.6435 & 31.7415 & 15.8666 & 70.3336 & 18.0401 & \textbf{5.7584}\\

F21 & Best & 1.2389e+03 & 1.0822e+03 & 1.0319e+03 & 500.3422 & 1.2062e+03 & \textbf{500.0000}\\
& Mean & 1.2549e+03 & 1.1096e+03 & 1.1535e+03 & 833.3210 & 1.2177e+03 & \textbf{711.0300}\\
& STD & 11.0959 & 12.0662 & 58.2933 & 337.3735 & 7.2659 & 275.1757\\

F22 & Best & 1.1649e+03 & 899.7449 & 1.0984e+03 & 1.0423e+03 & 1.0823e+03 & \textbf{581.3024}\\
& Mean & 1.2124e+03 & 948.8016 & 1.1922e+03 & 1.0992e+03 & 1.1153e+03 & \textbf{804.6198}\\
& STD & 29.1641 & 27.8554 & 61.1247 & 45.6596 & 27.1676 & 94.9291\\

F23 & Best & 1.2196e+03 & 1.1003e+03 & 1.1685e+03 & 550.6398 & 1.2005e+03 & \textbf{534.1640}\\
& Mean & 1.2527e+03 & 1.1165e+03 & 1.2037e+03 & 1.0637e+03 & 1.2174e+03 & \textbf{683.7794} \\
& STD & 19.0620 & 10.8319 & 18.6563 & 286.9826 & 15.2362 & 243.7339\\

F24 & Best & 1.2495e+03 & 937.2971 & 1.3316e+03 & 200.4646 & 1.2482e+03 & 323.1548\\
& Mean & 1.2795e+03 & 985.7198 & 1.4003e+03 & 638.5843 & 1.4003e+03 & \textbf{607.3364}\\
& STD & 15.3617 & 91.8088 & 33.5210 & 600.2639 & 33.5210 & 196.7787\\

F25 & Best & 1.7385e+03 & 1.9230e+03 & 1.7753e+03 & 1.6282e+03 & 1.7753e+03 & \textbf{1.3589e+03}\\
& Mean & 1.7848e+03 & 1.9684e+03 & 1.8208e+03 & 1.6326e+03 & 1.8208e+03 & \textbf{1.5204e+03}
\\
& STD & 27.3936 & 22.3652 & 27.7505 & 2.6468 & 27.7505 & 82.2932\\

\hline

\end{tabular}
\end{table*}

\begin{table*}
\fontsize{8pt}{12pt}
\selectfont

\centering
\caption{ Statistical analysis (Wilcoxon rank-sum test) on hybrid composite test functions with 30 dimensions}
\label{tab:wilcoxoncompos}
\resizebox{0.8\hsize}{!}{$%
\begin{tabular}{ll@{\qquad}llll@{\qquad}llll@{\qquad}llllllll@{\qquad}}

\toprule {Function No.}
\multirow{16}{*}{\raisebox{-\heavyrulewidth}} & \multicolumn{4}{c}{ GTBOA vs GA} & \multicolumn{4}{c} \ \ \ { GTBOA vs PSO} & \multicolumn{4}{c} \ \ \ {GTBOA vs ABC } & \multicolumn{4}{c} \ \ \ { } \\
\cmidrule{3-5}
\cmidrule{7-9}
\cmidrule{10-13}
\centering
& & pval & T+ & T-& winner & pval & T+ & T-& winner& pval & T+ & T-& winner \\
\midrule
F15 & & 7.8604e-70$(<0.05^+)$ & 725 & 265 & + & 4.0660e-149$(<0.05^+)$ & 863 & 127 & +& 4.2081e-131$(<0.05^+)$ & 870 & 120 & + \\
F16 & & 3.0334e-246$(<0.05^+)$ & 939 &51 & + & 1.0285e-248$(<0.05^+)$ & 939 & 50 & + & 1.7892e-310$(<0.05^+)$ & 990 & 0 & +\\
F17 & & 1.3046e-142$(<0.05^+)$ & 836 & 154 & + & 1.7612e-198$(<0.05^+)$ & 936 & 54 & +& 3.1198e-314$(<0.05^+)$ & 990 & 0 & + \\
F18 & & 1.2556e-312$(<0.05^+)$ & 988 & 2 & + & 1.5981e-203$(<0.05^+)$ & 888 & 102 & +& 5.3276e-284$(<0.05^+)$ & 990 & 0 & + \\
F19 & & 2.8290e-319$(<0.05^+)$ & 987 & 3 & + & 2.0457e-223$(<0.05^+)$ & 900 & 90 & +& 1.5210e-316$(<0.05^+)$ & 987 & 3 & + \\
F20 & & 2.1312e-286$(<0.05^+)$ & 973 & 17 & + & 1.3642e-176$(<0.05^+)$ & 850 & 140 & +& 2.6859e-270$(<0.05^+)$ & 990 & 0 & + \\
F21 & & 2.8565e-288$(<0.05^+)$ & 959 & 31 & + & 1.4215e-246$(<0.05^+)$ & 925 & 65 & +& 8.5974e-264$(<0.05^+)$ & 989 & 1 & + \\
F22 & & 3.7825e-319$(<0.05^+)$ & 984 & 6 & + & 3.5128e-47$(<0.05^+)$ & 691 & 299 & +& 1.0398e-201$(<0.05^+)$ & 964 & 26 & + \\
F23 & & 1.7376e-289$(<0.05^+)$ & 952 & 38 & + & 1.1670e-165$(<0.05^+)$ & 844 & 146 & +& 9.2559e-276$(<0.05^+)$ & 987 & 3 & + \\
F24 & & 2.1571e-291$(<0.05^+)$ & 967 & 23 & + & 1.1220e-168$(<0.05^+)$ & 901 & 89 & +& 1.0542e-311$(<0.05^+)$ & 990 & 0 & + \\
F25 & & 4.3952e-320$(<0.05^+)$ & 990 & 0 & + & 2.7120e-320$(<0.05^+)$ & 990 & 0 & +& 5.3673e-320$(<0.05^+)$ & 990 & 0 & + \\
\\
\hline
\\
\multirow{16}{*}{\raisebox{-\heavyrulewidth}} & \multicolumn{4}{c}{ GTBOA vs ALO} & \multicolumn{4}{c} \ \ \ { GTBOA vs BWO} & \multicolumn{4}{c} \ \ \ { } & \multicolumn{4}{c} \ \ \ { } \\
\cmidrule{3-5}
\cmidrule{7-9}

\centering
& & pval & T+ & T-& winner & pval & T+ & T-& winner \\
\\
F15 & & 1.3780e-166$(<0.05^+)$ & 879 & 111 & + & 1.7310e-176$(<0.05^+)$ & 839 & 151 & + \\
F16 & & 0.0218$(<0.05^+)$ & 533 & 457 & + & 1.7374e-264$(<0.05^+)$ & 919 & 71 & + \\
F17 & & 4.5041e-89$(<0.05^+)$ & 976 & 14 & + & 5.9849e-249$(<0.05^+)$ & 908 & 82 & + \\
F18 & & 1.2325e-219$(<0.05^+)$ & 987 & 3 & + & 1.4120e-312$(<0.05^+)$ & 988 & 2 & + \\
F19 & & 5.8540e-237$(<0.05^+)$ & 988 & 2 & + & 3.4180e-231$(<0.05^+)$ & 986 & 4 & + \\
F20 & & 0.1240$(>0.05^+)$ & 252 & 738 & - & 2.2213e-302$(<0.05^+)$ & 974 & 16 & + \\
F21 & & 1.4823e-81$(<0.05^+)$ & 959 & 31 & + & 1.6797e-305$(<0.05^+)$ & 953 & 37 & + \\
F22 & & 5.1015e-172$(<0.05^+)$ & 942 & 48 & + & 9.5479e-183$(<0.05^+)$ & 886 & 104 & + \\
F23 & & 6.6032e-50$(<0.05^+)$ & 789 & 201 & + & 9.4658e-293$(<0.05^+)$ & 943 & 47 & + \\
F24 & & 0.2153$(>0.05^+)$ & 168 & 822 & - & 5.6338e-314$(<0.05^+)$ & 965 & 25 & + \\
F25 & & 5.7644e-171$(<0.05^+)$ & 860 & 130 & + & 5.8710e-311$(<0.05^+)$ & 987 & 3 & + \\
\\

\bottomrule
\end{tabular}
$%
}%
\end{table*}

\begin{figure*}[htp!]
\centering
\includegraphics[scale=0.4]{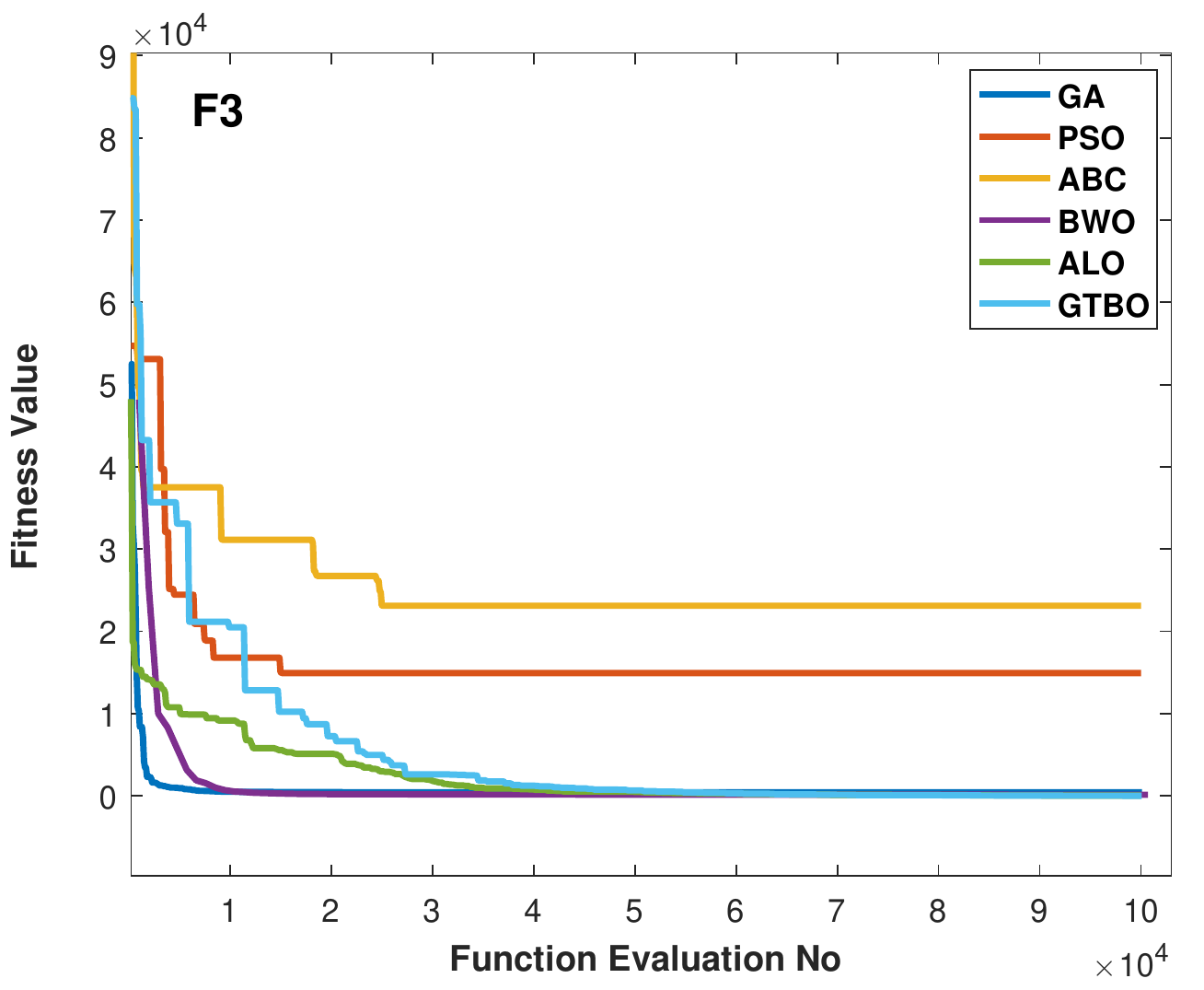}
\includegraphics[scale=0.4]{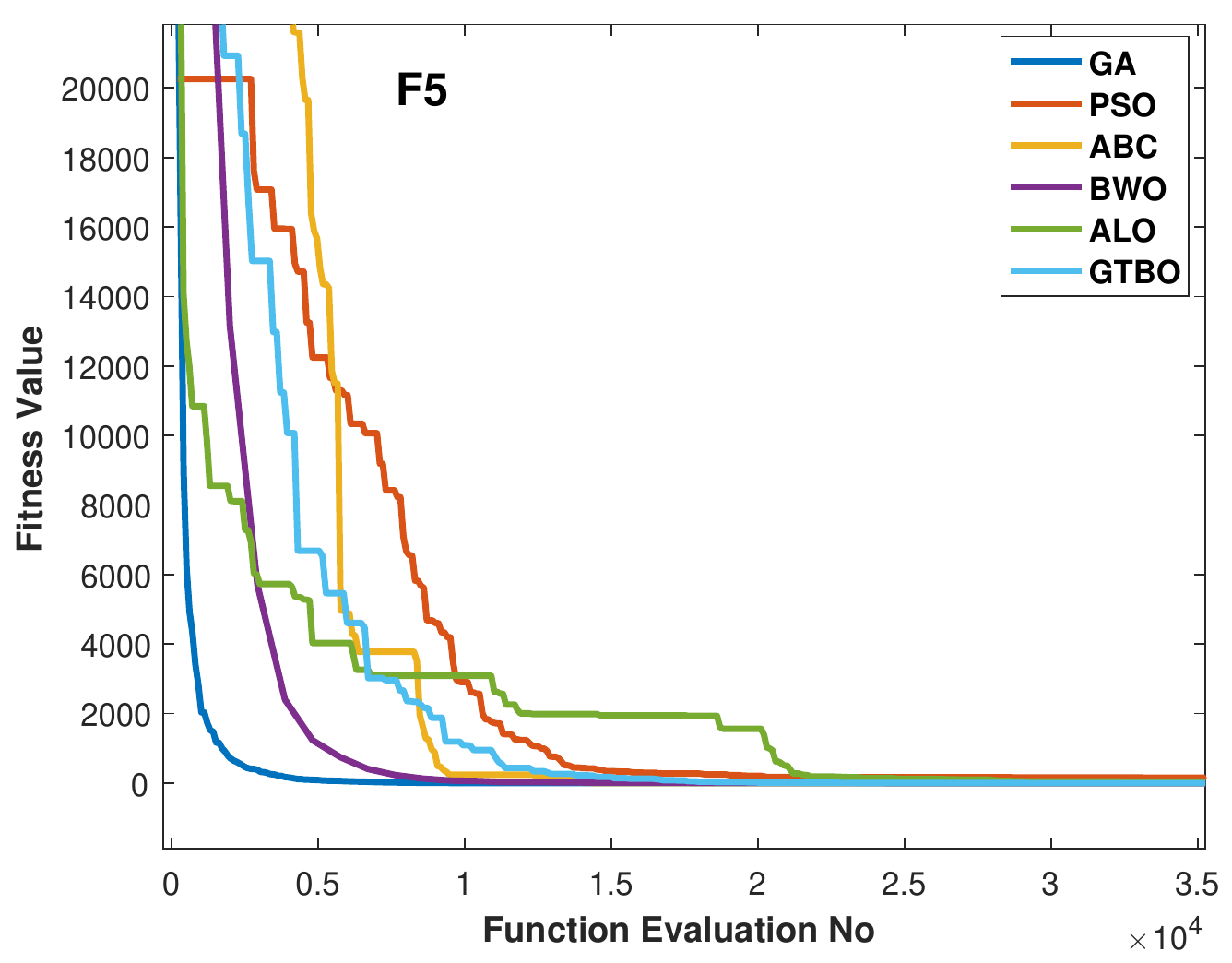}
\includegraphics[scale=0.4]{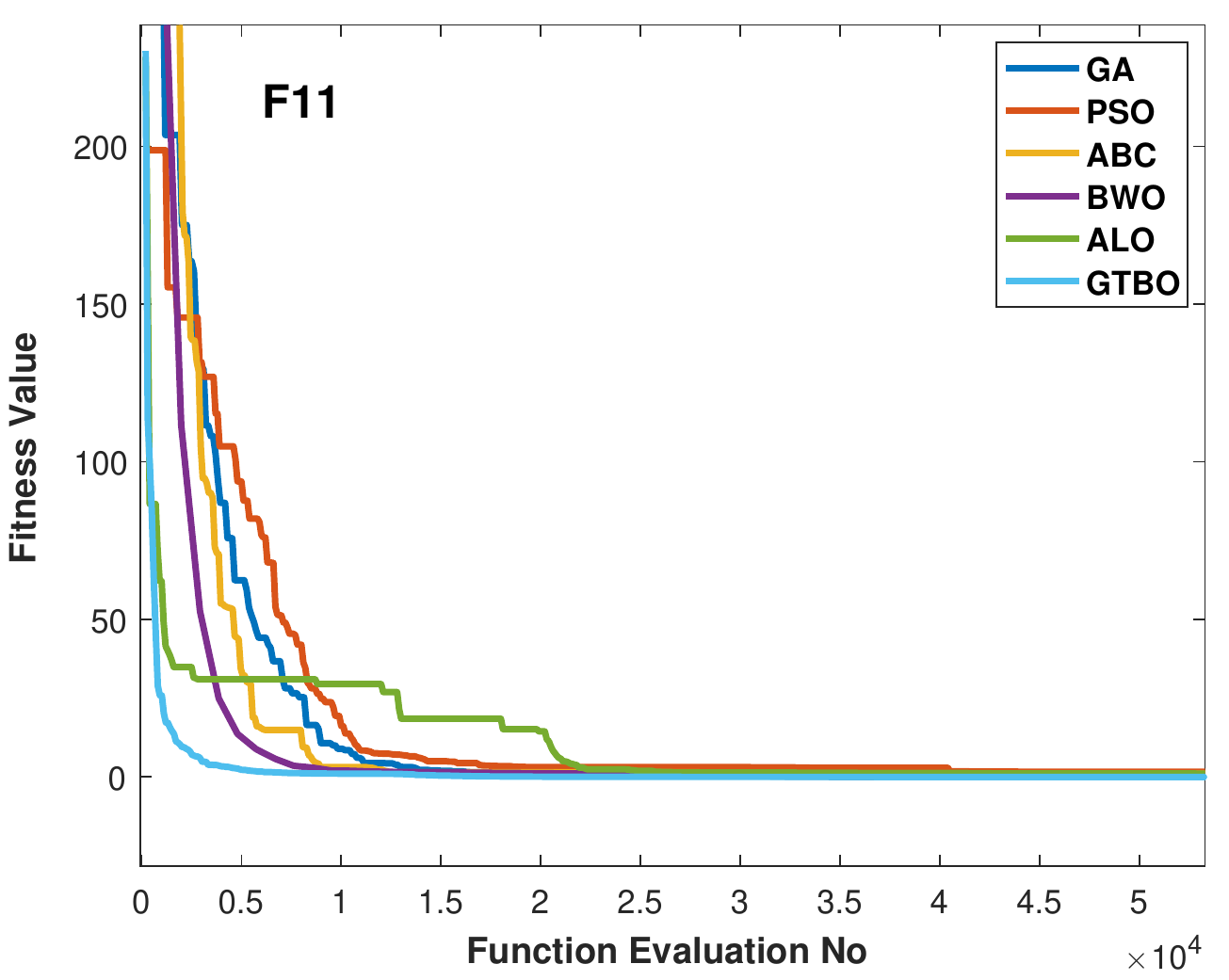}
\includegraphics[scale=0.4]{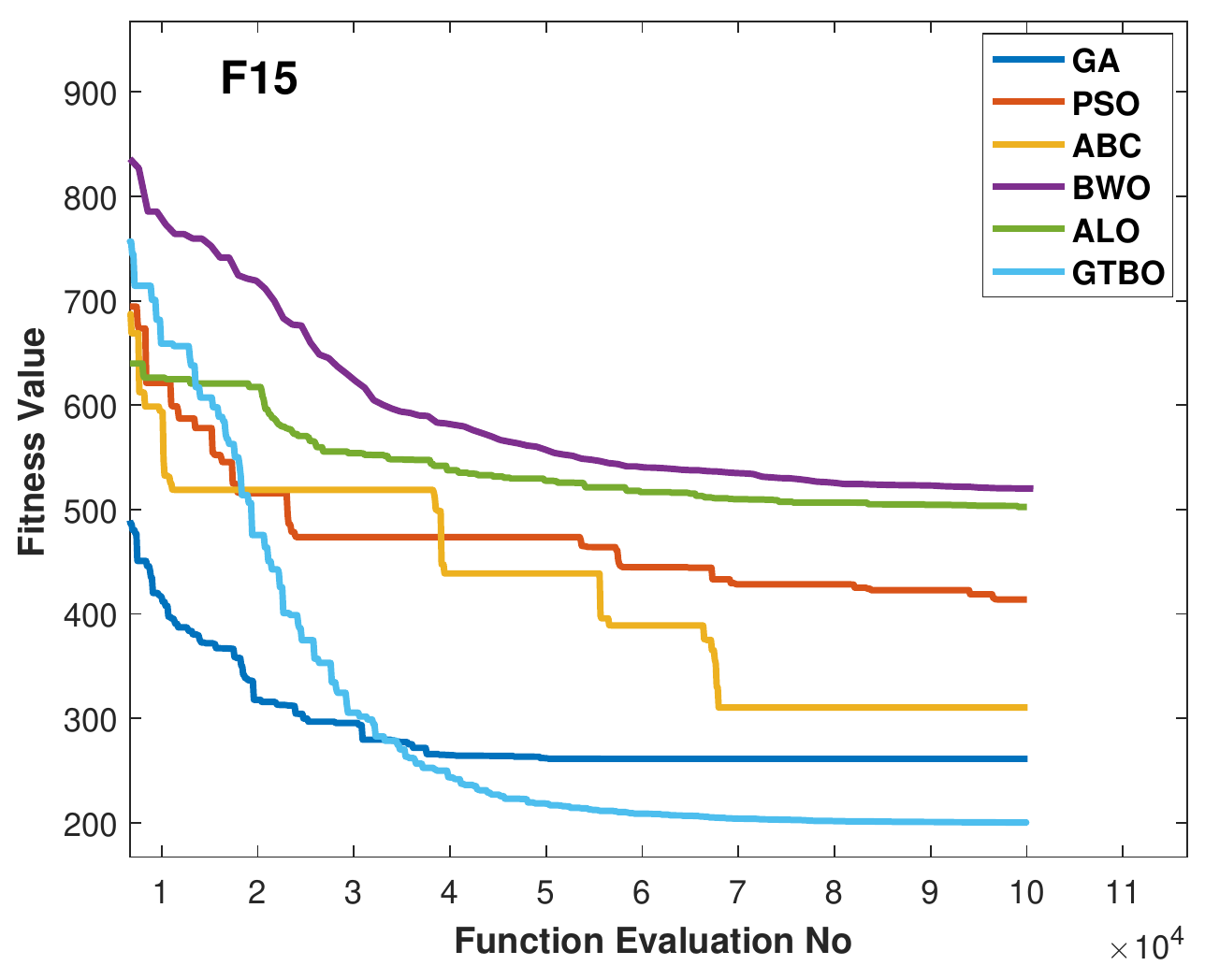}
\includegraphics[scale=0.4]{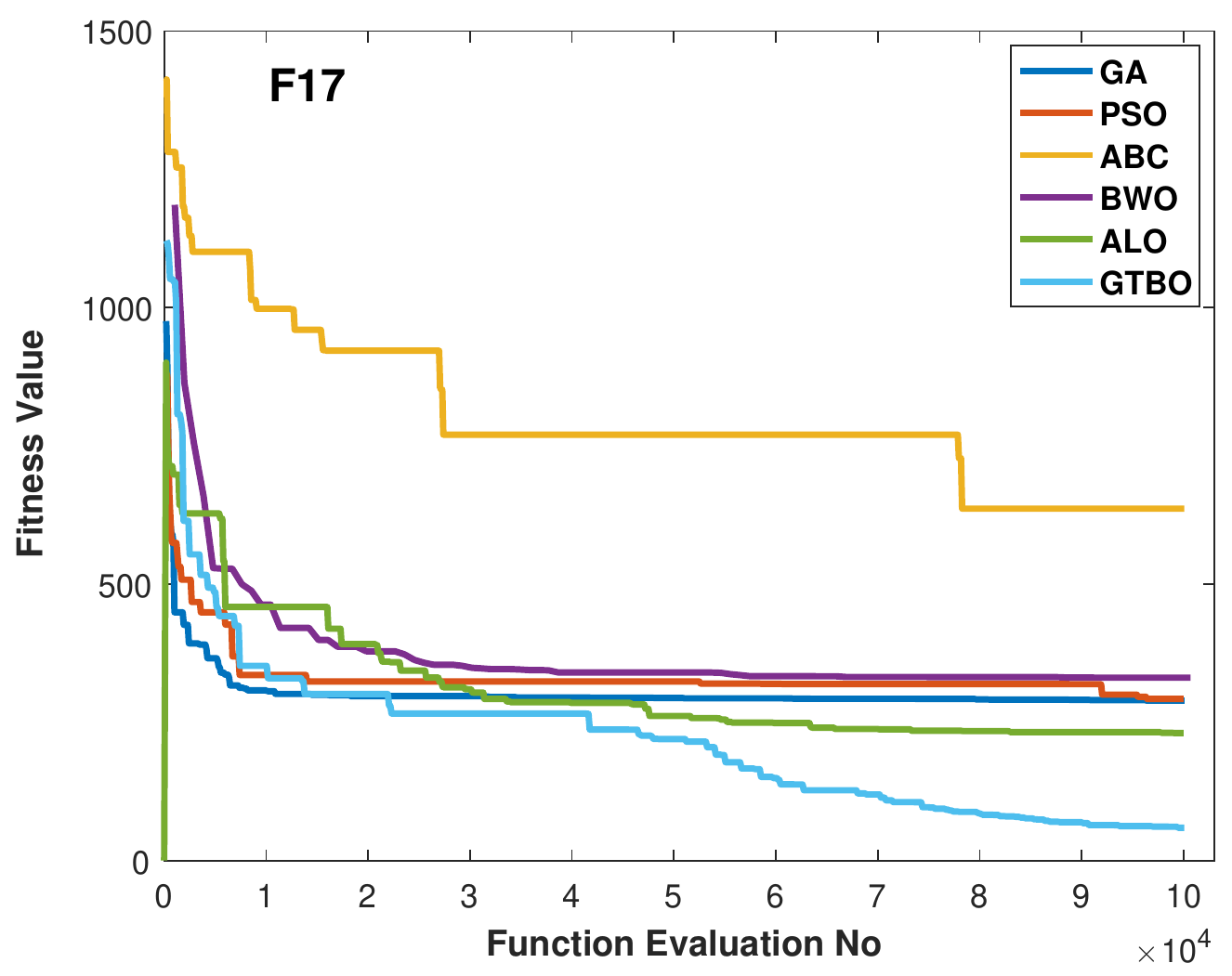}
\includegraphics[scale=0.4]{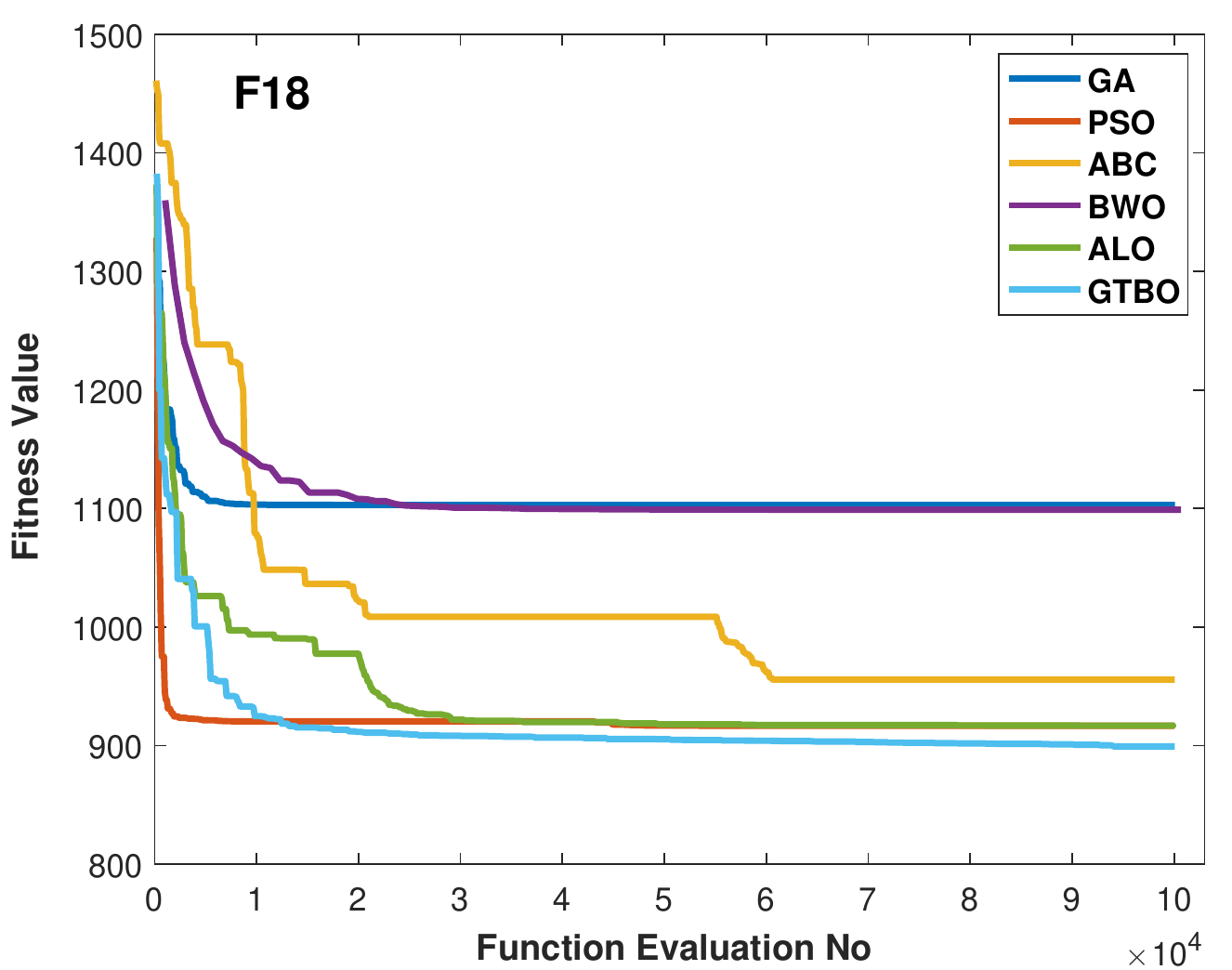}
\includegraphics[scale=0.4]{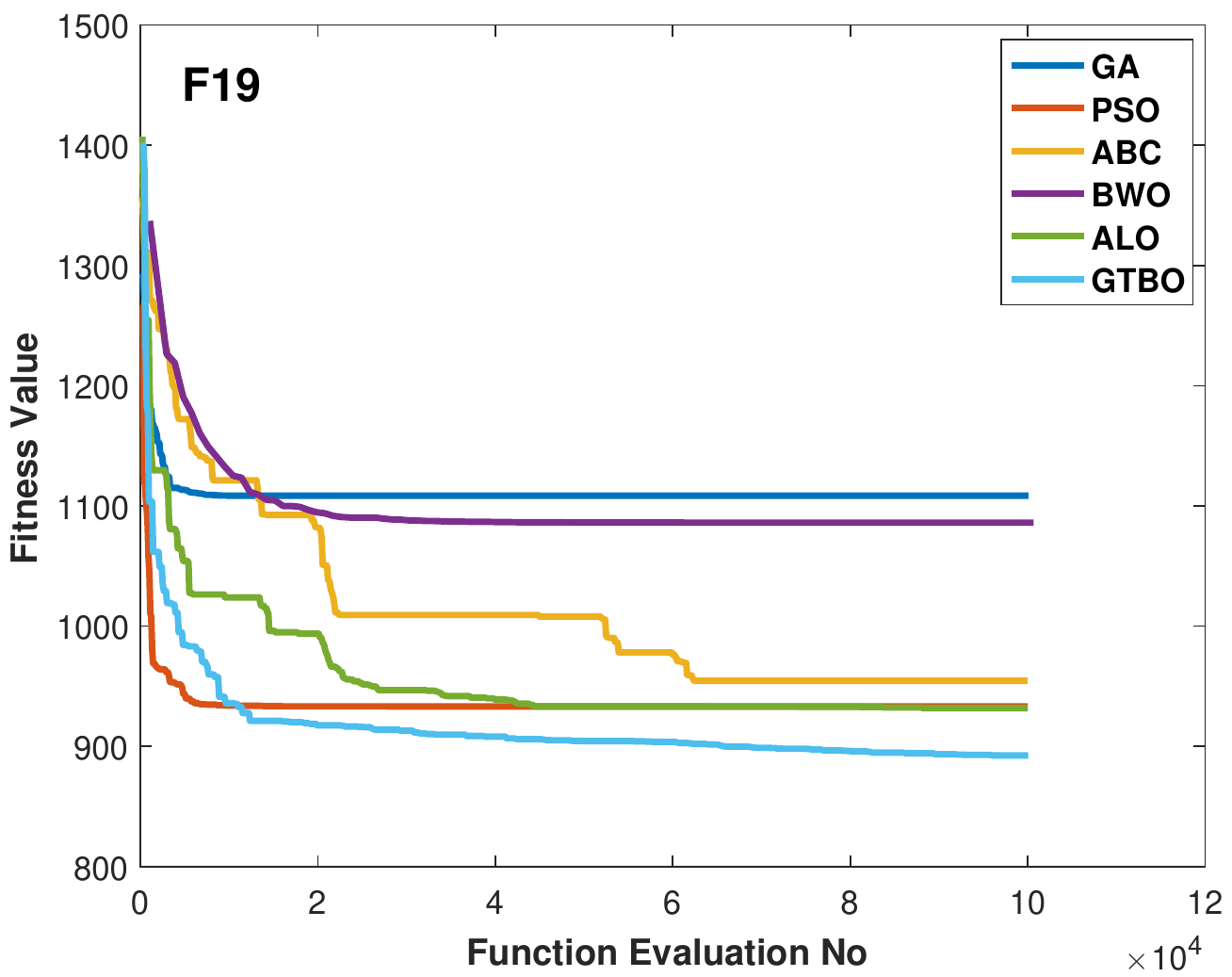}
\includegraphics[scale=0.4]{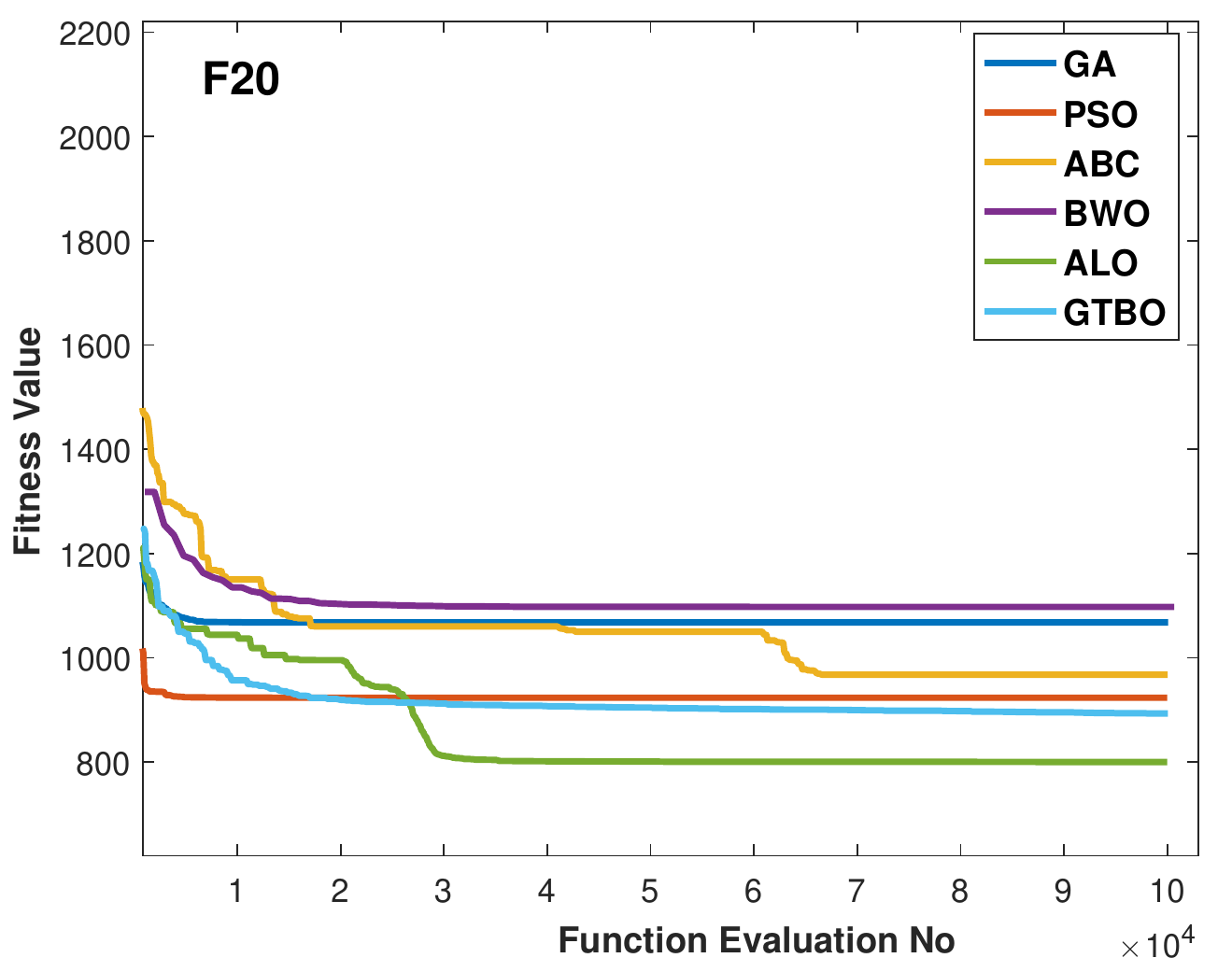}
\includegraphics[scale=0.4]{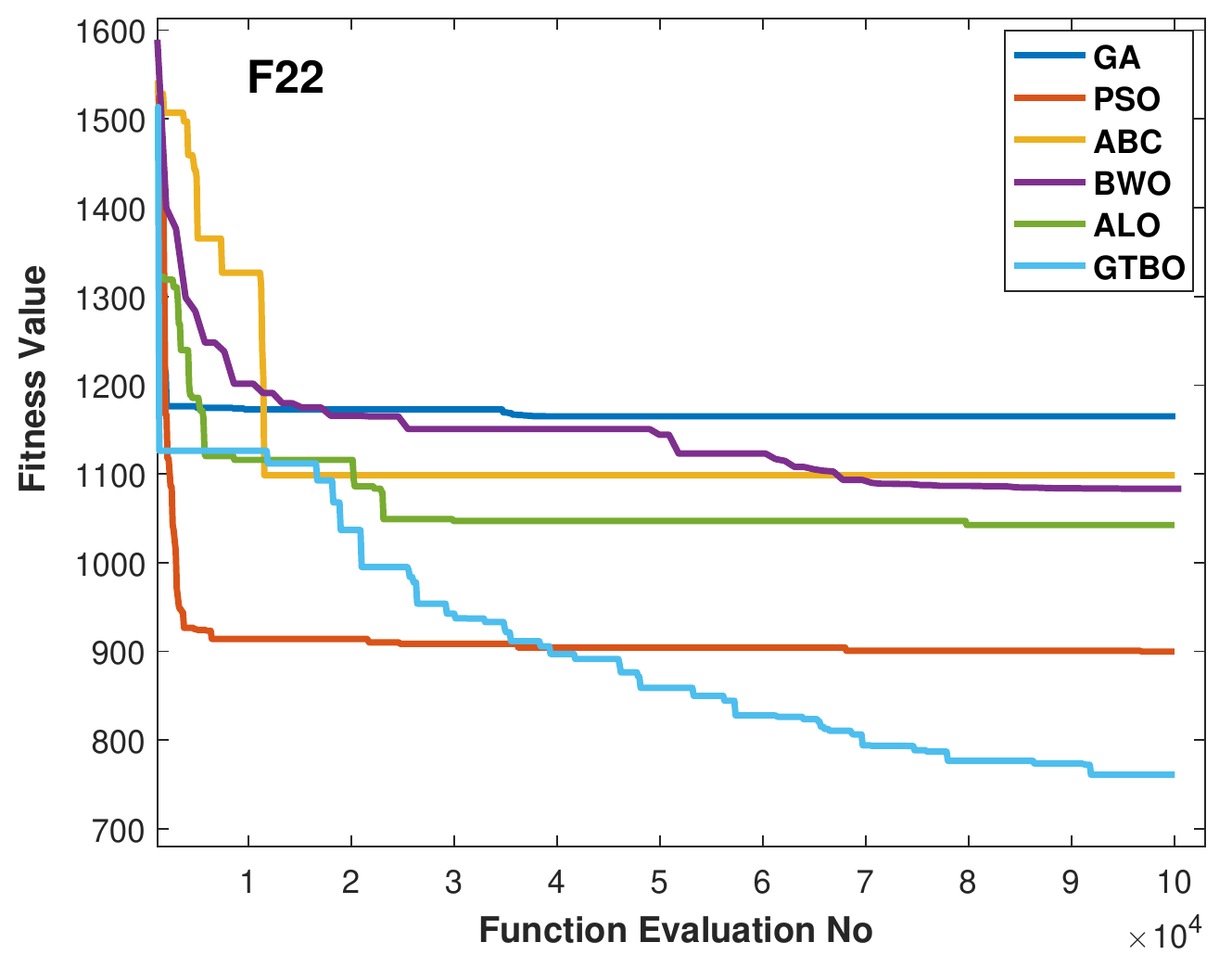}

\caption{Convergence curves for all the employed algorithms.}
\label{fig:gtboacurve}
\end{figure*}

Table~\ref{tab:uni1} shows the results of unimodal test functions using 30 dimensions. The GTBO reveals decent results on F1, F2, F3, F5, and F6. In these test functions, GTBO results are superior in terms of best, mean, and standard deviation. Other algorithms such as GA and BWO are in the next ranks. Regarding F4 and F7, GA achieved the best results. As mentioned earlier, unimodal test functions evaluate the algorithms in terms of exploitation and convergence rate. In this regard, as the algorithm demonstrates superior results in terms of unimodal test functions, it can be concluded that the GTBO has a reasonable efficiency in terms of exploitation and convergence speed. Figures~\ref{fig:gtboacurve} (a) and (b) show the convergence curve for function F3 and F5 which prove the efficiency of the proposed approach in converging to the global minimum. The algorithm's superiority is due to its employed operators that underpin a strong search mechanism resulting in optimal results.

It is not enough to prove the efficiency of the proposed algorithm just by showing the achieved results of the fitness function. The results should be tested statistically compared to the baseline algorithms (GA, PSO, ABC, ALO, and BWO) to demonstrate that it has not achieved the results by chance. In this regard, the Wilcoxon signed rank-sum test is employed to test the GTBO statistically at $5\%$ significance level~\cite{wilcoxon1992individual}. Table~\ref{tab:wilcoxonuni1} shows the results of GTBO as compared to the baseline algorithms where `$T+$' shows the sum of ranks for the GTBO, `$T-$' shows the sum of the rank for the baseline algorithms and $pval$ denotes the P-value achieved for each algorithm. There are two important concepts in the Wilcoxon test denoted as null and alternative hypothesis where the former shows that there is no significant difference between two existing algorithms while the latter denotes the existence of a significant difference between two algorithms. In Table~\ref{tab:wilcoxonuni1}, when the winner is shown by `+', it means that the GTBO is superior compared to its counterpart as the null hypothesis is rejected at $5\%$ significance level, however, when the winner is shown by `-' sign, it means that the null hypothesis is not rejected and there is no significant difference. According to Table~\ref{tab:wilcoxonuni1}, in most cases the GTBO is statistically superior against baseline algorithms.

The proposed approach has been applied to unimodal benchmark functions with higher dimensions too. Table~\ref{tab:uni2} shows the results of GTBO when the dimension number is 50. In this regard, the GTBO yields superior results in 4 (F1, F2, F5, and F6) out of the 7 test functions. The achieved results show that the proposed algorithm can still produce superior results when the number of dimensions increases significantly.

\subsection{Results on multimodal test functions}
The proposed algorithm is applied to some multimodal benchmark test functions and the results are shown in Tables~\ref{tab:multi1}, ~\ref{tab:wilcoxonmulti}, and ~\ref{tab:multi2}. According to Table~\ref{tab:multi1}, the GTBO shows superior results with respect to baseline algorithms when the dimension is 30. The proposed algorithm is superior in the majority of multimodal functions including F8, F10, F11, F12, and F13. The only exception is F9 where the GA algorithm has achieved better results than other algorithms. In terms of best-achieved results, the GA also shows a decent efficiency in F13 while the GTBO is in the second rank. A multimodal test function consists of more than one optimum and assesses the efficiency of the algorithm with regard to exploration and the local minimum avoidance. Due to the fact that the GTBO has revealed promising results in this type of test functions, it can be concluded that the proposed algorithm is efficient in terms of both exploration and local minima avoidance.

Similar to the unimodal test functions, the efficiency of the GTBO is also evaluated by employing the Wilcoxon signed rank-sum test to show that the proposed algorithm is statistically superior. In this regard, Table~\ref{tab:wilcoxonmulti} shows that the proposed algorithm is statistically superior compared to the baseline methods. According to Table~\ref{tab:wilcoxonmulti}, the GTBO is the winner in the majority of the cases and the only exceptions are F9 and F12. 

The results of the GTBO for multimodal functions are also tested on higher dimensions where Table~\ref{tab:multi2} shows that the proposed algorithm yields reasonable results when the dimension increases significantly. The proposed algorithm outperforms the baseline algorithms on majority of functions including F10, F11, F12, and F13. The poor performance of some baseline algorithms in multimodal functions reveals the reality that some of these algorithms are not efficient in terms of avoiding local minima to reach the global minimum. Another important factor is the exploration of these algorithms as the algorithms with poor exploration cannot explore the search space effectively.

\subsection{Results on hybrid composite test functions}
The proposed algorithm is applied to some composite functions and the results are shown in Table~\ref{tab:compos1}. These functions are challenging as they need to evaluate the performance of the algorithms in terms of the trade-off between exploration and exploitation. According to Table~\ref{tab:compos1}, the GTBO yields superior results in 10 out of 11 functions including F15, F16, F17, F18, F19, F21, F22, F23, F24, and F25. The proposed approach reveals a good efficiency in terms of convergence rate too. According to Figure~\ref{fig:gtboacurve}, the proposed algorithm converges faster to the global optima in most cases compared to the baseline algorithms.

Like other classes of test functions, the efficiency of the proposed algorithm is assessed statistically using Wilcoxon rank-sum test and the results are shown in Table~\ref{tab:wilcoxoncompos}. It is clear that the proposed algorithm demonstrates outstanding results in 9 out of 11 test functions, proving GTBO is statistically superior.

In this section, GTBO's performance is evaluated using different benchmark functions against a variety of baseline algorithms and the results reveal that the proposed algorithm is superior against baseline algorithms for most of the benchmark functions in terms of exploration, exploitation, avoidance of local minima, and balance between exploration and exploitation. In the GTBO, the employed color switching and survival operators play pivotal roles in creating good balance between exploration and exploitation resulting in decent performance. The color switching operator assists the algorithm in achieving global optima, while the survival operator helps the algorithm perform a better local search. The stopping criterion in the study is the number of function evaluations (NFE) which to the best of our knowledge is a fair stopping criterion. The poor performance of some algorithms lies in the concept of trapping into local minima where they cannot explore and exploit the search space effectively. In other words, they cannot make decent balance between exploration and exploitation and hence they cannot find the global minima.

\section{Application of GTBO to Engineering Problems}
\label{se:engopt}
This section shows the performance of applying the GTBO to two real problems in engineering. Both are real classical engineering design problems. The first one is called the welded beam design problem and the second one is called gear train design problem~\cite{mirjalili2016multi}. For the problems that have constraints, the simplest constraint handler (death penalty) is employed. Other constraint handlers such as special operators, repair algorithms, and hybrid methods are also used in the literature~\cite{coello2002theoretical, pasandideh2015optimization, jalali2016optimizing}. In the case of minimization problems when the agents violate the constraints, the penalty function is used to penalize the objective function resulting in larger fitness values. In this section, the only optimum values of the algorithm through 10 independent runs are presented and the statistical test is not performed as the statistical significance of the proposed algorithm against the same set of baseline algorithms was proved in the earlier experiments.

\subsection{Welded beam design problem}
\begin{figure}[htp!]
\centering
\includegraphics[scale=0.5]{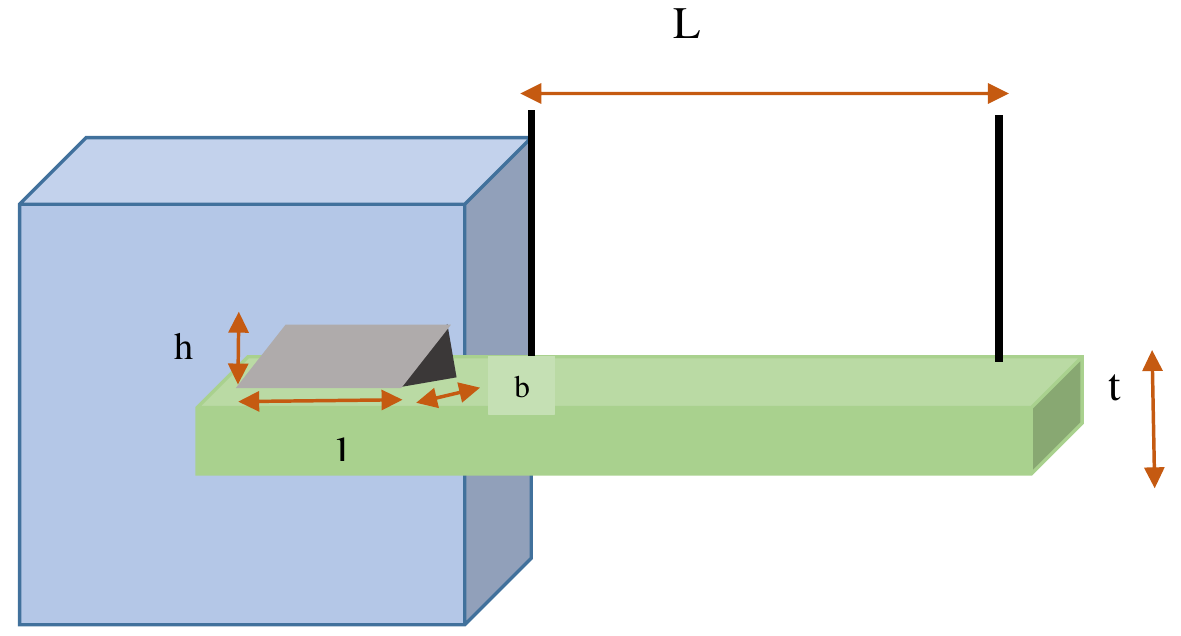}

\caption{Welded beam design parameters. }
\label{fig:weldbeam}
\end{figure}

The welded beam design problem is a minimization problem consisting of four variables namely, the length of bar attached to the weld (l), the bar height (t), the thickness of the weld (h), and the bar thickness (b)~\cite{sayed2018new, mirjalili2016multi, ragsdell1976optimal}. This problem is illustrated graphically by Figure~\ref{fig:weldbeam} where it also shows the structural parameters.

In addition, this problem has some defined constraints such as the beam stress ($\alpha$), the deflection of the beam ($\beta$), the bucking load on the bar (BL), the beam end deflection ($\gamma$), and the side constraints. The problem is defined using Equations~\ref{eq:weld1} and \ref{eq:weld2}.

\begin{equation}
\label{eq:weld1}
\resizebox{1\hsize}{!}{$%
\begin{cases}
Minimize \ \ f(\overrightarrow{z}) =1.10471z_1^2+0.04811z_3z_4(14.0+z_2),\\
\overrightarrow{z}=[z_1z_2z_3]=[hltb], \\
\\
Subject \ \ to \ \ \\
h_1(\overrightarrow{z})=\gamma(\overrightarrow{z})-\gamma_{max} \leq 0,\\
h_2(\overrightarrow{z})=\alpha(\overrightarrow{z})-\alpha_{max} \leq 0,\\
h_3(\overrightarrow{z})=z_1-z_4 \leq 0,\\
h_4(\overrightarrow{z})=0.10471z_1^2+0.04811z_3z_4(14.0+z_2)-5.0 \leq 0,\\
h_5(\overrightarrow{z})=0.125-z_1 \leq 0,\\
h_6(\overrightarrow{z})=\gamma(\overrightarrow{z})-\gamma_{max} \leq 0,\\
h_7(\overrightarrow{z})=B-BL(\overrightarrow{z}) \leq 0,\\
\\
where \\

0.1 \leq z_1 \leq 2,\\
0.1 \leq z_2 \leq 10,\\
0.1 \leq z_3 \leq 10,\\
0.1 \leq z_4 \leq 10,\\

\end{cases}
$%
}%
\end{equation}

Different values of parameters defined in Equation~\ref{eq:weld1} are declared in Equation~\ref{eq:weld2} as follows:
\begin{equation}
\label{eq:weld2}
\resizebox{1\hsize}{!}{$%
\begin{cases}
\gamma(\overrightarrow{z})=\sqrt{(\gamma')^2+2\gamma' \gamma^n \frac{z_2}{2R}+\gamma^{n2}}, \\
\gamma'=\frac{B}{\sqrt{2}z_1z_2}\\
\gamma^n=\frac{MR}{J}\\
M=B(L+\frac{z_2}{2})\\
R=\sqrt{\frac{z_2^2}{4}+(\frac{z_1+z_3}{2})^2}\\
J=2\bigg(\sqrt{2}z_1z_2 \bigg[{\frac{z_2^2}{4} \bigg(\frac{z_1+z_3}{2} \bigg)^2}\bigg]\bigg)\\
\alpha(\overrightarrow{z})=\frac{6PL}{Z_4Z_3^2}, \beta({\overrightarrow{z}})=\frac{4PL^3}{Ez_3^2z+z_4}\\
BL(\overrightarrow{z})=\frac{4.013E\sqrt{\frac{z_3^2z_4^6}{36}}}{L^2} \bigg(1-\frac{z_3}{2L}\sqrt{\frac{E}{4G}}\bigg)\\
P=6000lb, L=14in., \beta_{max}=0.25in., E=30 \times 10^6psi,\\
G=12 \times 10^6 psi, \gamma_{max}=13600psi, \alpha_{max}=30000psi.

\end{cases}
$%
}%
\end{equation}

The goal of this problem is to minimize the fabrication cost in the welded beam design consisting of seven constraints. It has a lot of applications in the design of machine elements and mechanical design problems~\cite{ragsdell1976optimal, mirjalili2016multi, he2004improved}. According to Table~\ref{tab:weldbeam}, the GTBO outperforms the baseline algorithms in terms of achieving the optimal cost as well as the set of optimal values for the four parameters, confirming that the algorithm can be used to minimize the fabrication cost of a reasonable welded beam design.

\begin{table}[H]
\fontsize{8pt}{12pt}
\selectfont
\centering
\caption{Results of welded beam design problem}
\label{tab:weldbeam}
\resizebox{1\hsize}{!}{$%
\begin{tabular}{ll@{\qquad}lll@{\qquad}lll@{\qquad}}

\toprule {Algorithms}
\multirow{7}{*}{\raisebox{-\heavyrulewidth}} & \multicolumn{4}{c}{ Welded beam optimal values for variables} \ \ \ \\
\cmidrule{2-6}

\centering

& & h \ \ \ \ \ \ \ \ \ \ \ \ \ &l \ \ \ \ \ \ \ \ \ \ \ \ \ & t \ \ \ \ \ \ \ \ \ \ \ \ \ & b \ \ \ \ \ \ \ \ \ & Optimal Value \\
\midrule
GA & & 0.3979 & 2.2478 & 6.0387 & 0.4652& 2.5890 \\
PSO & & 0.2637 & 4.6808 & 7.2402 & 0.3497& 2.6349 \\
ABC & & 0.1666 & 4.2434 & 9.1902 & 0.2055& 1.7880 \\
ALO & & 0.2049 & 3.2735 & 9.0382 & 0.2057& 1.6971 \\
BWO & & 0.1865 & 4.0128 & 8.3204 & 0.2427& 1.9039 \\
\textbf{GTBO} & & \textbf{0.2057} & \textbf{3.2530} & \textbf{9.0366} & \textbf{0.2057} & \textbf{1.6952} \\
\\

\bottomrule
\end{tabular}
$%
}%
\end{table}

\subsection{Gear train design problem}
The gear train design problem is defined using Equation~\ref{eq:gear1} as follows:

\begin{equation}
\label{eq:gear1}
\resizebox{0.7\hsize}{!}{$%
\begin{cases}
Let's \ \ consider: (\overrightarrow{n}) =[n_1,n_2,n_3,n_4], \\
Minimize: \ \ (\overrightarrow{n}) =\bigg(\frac{1}{6.931}-\frac{n_3n_2}{n_1n_4} \bigg)^2,\\

\\
Subject \ \ to: \ \ \\
12 \leq n_1,n_2,n_3,n_4 \leq 60\\

\end{cases}
$%
}%
\end{equation}

\begin{figure}[htp!]
\centering
\includegraphics[scale=0.5]{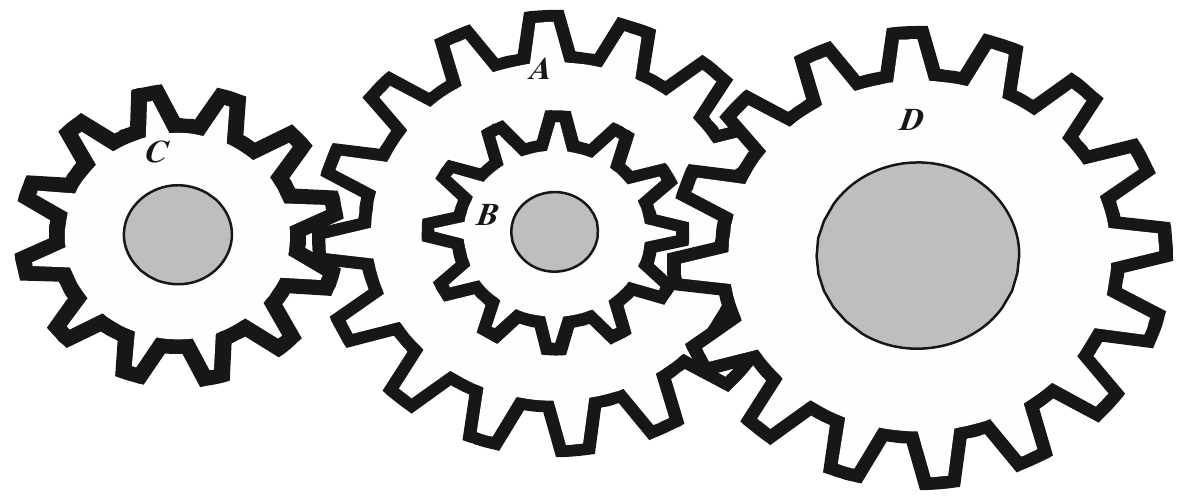}

\caption{Gear train design problem parameters~\cite{mirjalili2015ant}. }
\label{fig:geartrain}
\end{figure}

The gear train design problem is a mechanical engineering problem aiming to minimize the gear ratio using four gears of a train~\cite{gandomi2014interior, sandgren1990nonlinear}. The number of teeth of the gears is considered as the parameters of the problem. This problem has no constraints, although the ranges of the variables can be considered as the constraints of the problem. In Figure~\ref{fig:geartrain}, different parameters of this problem are depicted. According to the results in Table~\ref{tab:geartrain}, the GTBO again outperforms the baseline algorithms in terms of obtaining a superior gear ratio and optimal values for the four parameters. It is worth noting that as the gear train design problem is a discrete problem, this experiment also confirms that GTBO is efficient in solving discrete problems.

\begin{table}[H]
\fontsize{8pt}{12pt}
\selectfont
\centering
\caption{Results of gear train design problem}
\label{tab:geartrain}
\resizebox{1\hsize}{!}{$%
\begin{tabular}{ll@{\qquad}lll@{\qquad}lll@{\qquad}}

\toprule {Algorithms}
\multirow{7}{*}{\raisebox{-\heavyrulewidth}} & \multicolumn{4}{c}{ Gear train optimal values for variables} \ \ \ \\
\cmidrule{2-6}

\centering

& & $n_A$ \ \ \ \ \ \ \ \ \ \ \ \ \ &$n_B$ \ \ \ \ \ \ \ \ \ \ \ \ \ & $n_C$\ \ \ \ \ \ \ \ \ \ \ \ \ & $n_D$ \ \ \ \ \ \ \ \ \ & Optimal Value \\
\midrule
GA & & 50 & 20 & 17 & 47 & 1.6160e-12 \\
PSO & & 28 & 12 & 16 & 49 & 7.2211e-06 \\
ABC & & 43 & 13 & 16 & 35 & 1.1096e-17 \\
ALO & & 58 & 16 & 29 & 59& 2.1771e-15 \\
BWO & & 50 & 18 & 17 & 46 & 7.5421e-17 \\
\textbf{GTBO} & & \textbf{51} & \textbf{32} & \textbf{12} & \textbf{52} & \textbf{2.0211e-20} \\
\\

\bottomrule
\end{tabular}
$%
}%
\end{table}

\subsection{Sensitivity analysis on key parameters}
This section discusses the sensitivity analysis of the key parameters in the GTBO in order to understand the impacts of different values of $M_r$ (mature rate) and $S_r$ (survival rate) on the overall algorithm performance. In addition, the two operators that generate solutions in the algorithm are analyzed to find out their performance and impact on the algorithm's efficiency.

\begin{table}[H]
\fontsize{10pt}{15pt}
\selectfont
\centering
\caption{Sensitivity analysis on mature rate ($M_r$)}
\label{tab:sensmature}
\resizebox{1\hsize}{!}{$%
\begin{tabular}{ll@{\qquad}lll@{\qquad}lll@{\qquad}}

\toprule {Function No.}
\multirow{7}{*}{\raisebox{-\heavyrulewidth}} & \multicolumn{4}{c}{ GTBO optimal values for different mature rates} \ \ \ \\
\cmidrule{2-6}

\centering

& & $0.2$ \ \ \ \ \ \ \ \ \ \ \ \ \ &$0.4$ \ \ \ \ \ \ \ \ \ \ \ \ \ & $0.6$\ \ \ \ \ \ \ \ \ \ \ \ \ & $0.8$ \ \ \ \ \ \ \ \ \ & 0.9 \\
\midrule
F2 & & 2.4604e-14 & 1.8303e-11 & 4.4458e-09 & 1.4349e-07 & 8.0541e-07 \\
F5 & & 19.7865 & 64.3749 & 34.1285 & 31.2356 & 35.8112 \\
F10 & & 4.7373e-12 & 4.9744e-10 & 2.3079e-08 & 5.6245e-07 & 2.0360e-06\\
F12 & & 0.5022 & 6.8447e-08 & 2.9187e-06 & 2.0542e-04 & 4.0006e-04 \\
F15 & & 421.0066 & 304.7715 & 400.0000 & 278.0010 & 285.8998 \\
F21 & & 511.0000 & 500.0000 & 500.0000 & 500.0000 & 500.000 \\
\\

\bottomrule
\end{tabular}
$%
}%
\end{table}

Regarding the key parameters, $M_r$ is tested using different values of (0.2, 0.4, 0.6, 0.8, 0.9) and $S_r$ is tested on the values of (0.2, 0.4, 0.6, 0.8). In every class of test function, two functions are selected randomly. The algorithm performed 30 independent runs and used 30 dimensions for unimodal and multimodal, as well as 10 dimensions for the composition functions. Table~\ref{tab:sensmature} shows the sensitivity analysis results on the mature rate ($M_r$) control parameter. It is clear that the GTBO has a better performance on unimodal and multimodal problems when $M_r$ is between [0.2, 0.4], while it has superior performance in composite problems when it has larger values such as 0.8 and 0.9.

\begin{table}[H]
\fontsize{10pt}{12pt}
\selectfont
\centering
\caption{Sensitivity analysis on survival rate ($S_r$)}
\label{tab:sensurvive}
\resizebox{1\hsize}{!}{$%
\begin{tabular}{ll@{\qquad}lll@{\qquad}lll@{\qquad}}

\toprule {Function No.}
\multirow{7}{*}{\raisebox{-\heavyrulewidth}} & \multicolumn{4}{c}{ GTBO optimal values for different survival rates} \ \ \ \\
\cmidrule{2-6}

\centering

& & $0.2$ \ \ \ \ \ \ \ \ \ \ \ \ \ &$0.4$ \ \ \ \ \ \ \ \ \ \ \ \ \ & $0.6$\ \ \ \ \ \ \ \ \ \ \ \ \ & $0.8$ \\
\midrule
F2 & & 1.4349e-07 & 1.5478e-07 & 4.4734e-07 & 2.1932e-06 \\
F5 & & 31.2356 & 61.6426 & 57.6874 & 90.6470 \\
F10 & & 5.6245e-07 & 1.4420e-06 & 2.7178e-06 & 1.5281e-05 \\
F12 & & 2.0542e-04 & 0.0524 & 0.2185 & 0.1459 \\
F15 & & 252.8611 & 235.5687 & 301.2868 & 261.9206 \\
F21 & & 500.0000 & 500.0000 & 586.9541 & 500.0000 \\
\\

\bottomrule
\end{tabular}
$%
}%
\end{table}

According to Table~\ref{tab:sensurvive}, the GTBO has superior results when the value of $S_r$ is initialized to 0.2. This study utilized this value in order to obtain the best results. In this test, the value of $M_r$ is initialized to 0.4 for all the cases.

Based on the results in Tables~\ref{tab:sensmature} and ~\ref{tab:sensurvive}, the optimal values of $M_r$ and $S_r$ are 0.4 and 0.2 respectively. 

\begin{table}[H]
\fontsize{10pt}{15pt}
\selectfont

\centering
\caption{ Sensitivity analysis on operators defined in GTBO}
\label{tab:sensoperator}
\resizebox{1\hsize}{!}{$%
\begin{tabular}{ll@{\qquad}llll}

\toprule {Function No.}
\multirow{1}{*}{} & \multicolumn{3}{c}{ Sensitivity analysis on GTBO operators} \\
\cmidrule{3-5}

\centering
& Criteria & Color Switching Operator & Survival Operator & Composition of Both Operators \\
\midrule

F2 & Best & 1.0000e-04 & 0.1779 & \textbf{6.4949e-08}\\
& Mean & 0.8523 &2.3908 & \textbf{1.4349e-07}\\
& STD & 2.5567 & 2.1834 & \textbf{6.4830e-08}\\

F5 & Best & 22.7657 & 2.5408e+04 & 24.3678\\
& Mean & 43.5455 & 4.2193e+05 & \textbf{31.2356} \\
& STD & 29.3606 & 3.2102e+05 & \textbf{17.3351} \\

F10 & Best & 2.0000e-04 & 5.4435 & \textbf{1.9400e-07} \\
& Mean & 5.4000e-04 & 7.2504 & \textbf{5.6245e-07}\\
& STD & 2.5473e-04 & 1.3102 & \textbf{3.6438e-07} \\

F12 & Best & 0.0065 & 1.9051 & \textbf{2.4174e-06} \\
& Mean & 0.1993 & 88.0779 & \textbf{2.0542e-04}\\
& STD & 0.2587 & 189.7275 & \textbf{4.9198e-04} \\

\bottomrule
\end{tabular}
$%
}%
\end{table}

Table~\ref{tab:sensoperator} shows the results of sensitivity analysis on different employed operators (color-switching operator, survival operator, and a composition of both) in GTBO. It has been done just for some randomly chosen unimodal and multimodal test functions over 30 independent runs. As it is shown in Table~\ref{tab:sensoperator}, the results using color-switching operator are more robust than the cases when the algorithm just uses survival operator. Furthermore, when both operators are used, the results are much more superior than only using one operator individually. It further confirms that while the color switching operator is superior in terms of global search and the survival operator excels in local search, the composition of the two operators not only improves the population diversity but also increases the search capability, resulting in superior balance between exploration and the exploitation.

\section{Conclusion and Future Works}
\label{lab:conclusion}
The main contribution of this work is a novel optimization algorithm inspired by the natural life of the golden tortoise beetle. The golden tortoise beetle uses a color-switching mechanism in order to mate and enhance its chance for reproduction. In addition, this beetle uses a kind of shield to protect its larvae against predators. These notions are mathematically modeled to design a novel optimization algorithm. The algorithm has been tested on 24 benchmark functions using different classes of unimodal, multimodal, and hybrid composite functions against some well-known baseline algorithms including GA, PSO, ABC, ALO, and BWO. The superiority of the algorithm on unimodal functions proves its efficiency in exploitation and convergence rate. In terms of multimodal, the algorithm also outperforms other algorithms, which shows the potential of the GTBO in exploration and the local minima avoidance. It can be concluded that the algorithm can make a trade-off between exploration and exploitation as it has achieved remarkable results in terms of hybrid composite functions. The algorithm is also tested statistically using Wilcoxon signed rank-sum test whose results have revealed that the algorithm is statistically superior with respect to the baseline algorithms.

The algorithm performance has also been tested on two well-known engineering problems and the results conclude that GTBO is efficient in solving real problems with constraints and unknown search spaces. A sensitivity analysis has also been performed to provide some evidence for choosing different values of the key control parameters and for combining the two operators to achieve a suitable trade-off between exploration and exploitation. Because of the random selection of some solutions in the algorithm, the GTBO has a high probability of escaping from the local minima. Moreover, population diversity plays a pivotal role in achieving the global minimum.

In future work, the GTBO will be applied to different optimization problems. The algorithm's key control parameters such as mature rate and the survival rate can be designed with an adaptive approach. Finally, a hybrid approach combining the GTBO with other optimizers can be considered for further performance improvement.

\bibliographystyle{elsarticle-num}

\bibliography{bibliography}

\begin{thebibliography}{10}
\expandafter\ifx\csname url\endcsname\relax
  \def\url#1{\texttt{#1}}\fi
\expandafter\ifx\csname urlprefix\endcsname\relax\def\urlprefix{URL }\fi
\expandafter\ifx\csname href\endcsname\relax
  \def\href#1#2{#2} \def\path#1{#1}\fi

\bibitem{yang2010nature}
X.-S. Yang, Nature-inspired metaheuristic algorithms, Luniver press, 2010.

\bibitem{rajabioun2011cuckoo}
R.~Rajabioun, Cuckoo optimization algorithm, Applied soft computing 11~(8)
  (2011) 5508--5518.

\bibitem{johnson2002class}
A.~W. Johnson, S.~H. Jacobson, A class of convergent generalized hill climbing
  algorithms, Applied Mathematics and Computation 125~(2-3) (2002) 359--373.

\bibitem{holland1992genetic}
J.~H. Holland, Genetic algorithms, Scientific american 267~(1) (1992) 66--73.

\bibitem{tarkhaneh2019adaptive}
O.~Tarkhaneh, H.~Shen, An adaptive differential evolution algorithm to optimal
  multi-level thresholding for mri brain image segmentation, Expert Systems
  with Applications 138 (2019) 112820.

\bibitem{hatamlou2013black}
A.~Hatamlou, Black hole: A new heuristic optimization approach for data
  clustering, Information sciences 222 (2013) 175--184.

\bibitem{zhang2020binary}
Y.~Zhang, D.-w. Gong, X.-z. Gao, T.~Tian, X.-y. Sun, Binary differential
  evolution with self-learning for multi-objective feature selection,
  Information Sciences 507 (2020) 67--85.

\bibitem{mafarja2017hybrid}
M.~M. Mafarja, S.~Mirjalili, Hybrid whale optimization algorithm with simulated
  annealing for feature selection, Neurocomputing 260 (2017) 302--312.

\bibitem{mirjalili2014grey}
S.~Mirjalili, S.~M. Mirjalili, A.~Lewis, Grey wolf optimizer, Advances in
  engineering software 69 (2014) 46--61.

\bibitem{mirjalili2015ant}
S.~Mirjalili, The ant lion optimizer, Advances in engineering software 83
  (2015) 80--98.

\bibitem{gandomi2013cuckoo}
A.~H. Gandomi, X.-S. Yang, A.~H. Alavi, Cuckoo search algorithm: a
  metaheuristic approach to solve structural optimization problems, Engineering
  with computers 29~(1) (2013) 17--35.

\bibitem{olorunda2008measuring}
O.~Olorunda, A.~P. Engelbrecht, Measuring exploration/exploitation in particle
  swarms using swarm diversity, in: 2008 IEEE congress on evolutionary
  computation (IEEE world congress on computational intelligence), IEEE, 2008,
  pp. 1128--1134.

\bibitem{alba2005exploration}
E.~Alba, B.~Dorronsoro, The exploration/exploitation tradeoff in dynamic
  cellular genetic algorithms, IEEE transactions on evolutionary computation
  9~(2) (2005) 126--142.

\bibitem{lin2009auto}
L.~Lin, M.~Gen, Auto-tuning strategy for evolutionary algorithms: balancing
  between exploration and exploitation, Soft Computing 13~(2) (2009) 157--168.

\bibitem{cagnina2008solving}
L.~C. Cagnina, S.~C. Esquivel, C.~A.~C. Coello, Solving engineering
  optimization problems with the simple constrained particle swarm optimizer,
  Informatica 32~(3).

\bibitem{yang2012bat}
X.-S. Yang, A.~H. Gandomi, Bat algorithm: a novel approach for global
  engineering optimization, Engineering computations.

\bibitem{papa2017quaternion}
J.~P. Papa, G.~H. Rosa, D.~R. Pereira, X.-S. Yang, Quaternion-based deep belief
  networks fine-tuning, Applied Soft Computing 60 (2017) 328--335.

\bibitem{de2018swarm}
F.~De~Rango, N.~Palmieri, X.-S. Yang, S.~Marano, Swarm robotics in wireless
  distributed protocol design for coordinating robots involved in cooperative
  tasks, Soft Computing 22~(13) (2018) 4251--4266.

\bibitem{mirjalili2016multi}
S.~Mirjalili, S.~M. Mirjalili, A.~Hatamlou, Multi-verse optimizer: a
  nature-inspired algorithm for global optimization, Neural Computing and
  Applications 27~(2) (2016) 495--513.

\bibitem{wolpert1997no}
D.~H. Wolpert, W.~G. Macready, No free lunch theorems for optimization, IEEE
  transactions on evolutionary computation 1~(1) (1997) 67--82.

\bibitem{digalakis2001benchmarking}
J.~G. Digalakis, K.~G. Margaritis, On benchmarking functions for genetic
  algorithms, International journal of computer mathematics 77~(4) (2001)
  481--506.

\bibitem{molga2005test}
M.~Molga, C.~Smutnicki, Test functions for optimization needs, Test functions
  for optimization needs 101.

\bibitem{suganthan2005problem}
P.~N. Suganthan, N.~Hansen, J.~J. Liang, K.~Deb, Y.-P. Chen, A.~Auger,
  S.~Tiwari, Problem definitions and evaluation criteria for the cec 2005
  special session on real-parameter optimization, KanGAL report 2005005~(2005)
  (2005) 2005.

\bibitem{wilcoxon1992individual}
F.~Wilcoxon, Individual comparisons by ranking methods, in: Breakthroughs in
  statistics, Springer, 1992, pp. 196--202.

\bibitem{yang2020nature}
X.-S. Yang, Nature-inspired optimization algorithms: challenges and open
  problems, Journal of Computational Science (2020) 101104.

\bibitem{mirjalili2016whale}
S.~Mirjalili, A.~Lewis, The whale optimization algorithm, Advances in
  engineering software 95 (2016) 51--67.

\bibitem{storn1997differential}
R.~Storn, K.~Price, Differential evolution--a simple and efficient heuristic
  for global optimization over continuous spaces, Journal of global
  optimization 11~(4) (1997) 341--359.

\bibitem{yao1999evolutionary}
X.~Yao, Y.~Liu, G.~Lin, Evolutionary programming made faster, IEEE Transactions
  on Evolutionary computation 3~(2) (1999) 82--102.

\bibitem{sprave1994linear}
J.~Sprave, Linear neighborhood evolution strategy, in: Proceedings of the 3rd
  Annual Conference on Evolutionary Programming, World Scientific, 1994, pp.
  42--51.

\bibitem{koza1992genetic}
J.~R. Koza, J.~R. Koza, Genetic programming: on the programming of computers by
  means of natural selection, Vol.~1, MIT press, 1992.

\bibitem{simon2008biogeography}
D.~Simon, Biogeography-based optimization, IEEE transactions on evolutionary
  computation 12~(6) (2008) 702--713.

\bibitem{rashedi2009gsa}
E.~Rashedi, H.~Nezamabadi-Pour, S.~Saryazdi, Gsa: a gravitational search
  algorithm, Information sciences 179~(13) (2009) 2232--2248.

\bibitem{kaveh2010novel}
A.~Kaveh, S.~Talatahari, A novel heuristic optimization method: charged system
  search, Acta Mechanica 213~(3-4) (2010) 267--289.

\bibitem{formato2007central}
R.~Formato, Central force optimization: a new metaheuristic with applications
  in applied electromagnetics. prog electromagn res 77: 425--491 (2007).

\bibitem{alatas2011acroa}
B.~Alatas, Acroa: artificial chemical reaction optimization algorithm for
  global optimization, Expert Systems with Applications 38~(10) (2011)
  13170--13180.

\bibitem{kaveh2012new}
A.~Kaveh, M.~Khayatazad, A new meta-heuristic method: ray optimization,
  Computers \& structures 112 (2012) 283--294.

\bibitem{shah2011principal}
H.~Shah-Hosseini, Principal components analysis by the galaxy-based search
  algorithm: a novel metaheuristic for continuous optimisation, International
  Journal of Computational Science and Engineering 6~(1-2) (2011) 132--140.

\bibitem{carvalho2007particle}
M.~Carvalho, T.~B. Ludermir, Particle swarm optimization of neural network
  architectures andweights, in: 7th International Conference on Hybrid
  Intelligent Systems (HIS 2007), IEEE, 2007, pp. 336--339.

\bibitem{dorigo2006ant}
M.~Dorigo, M.~Birattari, T.~Stutzle, Ant colony optimization, IEEE
  computational intelligence magazine 1~(4) (2006) 28--39.

\bibitem{karaboga2007powerful}
D.~Karaboga, B.~Basturk, A powerful and efficient algorithm for numerical
  function optimization: artificial bee colony (abc) algorithm, Journal of
  global optimization 39~(3) (2007) 459--471.

\bibitem{hayyolalam2020black}
V.~Hayyolalam, A.~A.~P. Kazem, Black widow optimization algorithm: A novel
  meta-heuristic approach for solving engineering optimization problems,
  Engineering Applications of Artificial Intelligence 87 (2020) 103249.

\bibitem{yang2009cuckoo}
X.-S. Yang, S.~Deb, Cuckoo search via l{\'e}vy flights, in: 2009 World congress
  on nature \& biologically inspired computing (NaBIC), IEEE, 2009, pp.
  210--214.

\bibitem{li2003new}
X.~Li, A new intelligent optimization-artificial fish swarm algorithm, Doctor
  thesis, Zhejiang University of Zhejiang, China (2003) 27.

\bibitem{pinto2007wasp}
P.~C. Pinto, T.~A. Runkler, J.~M. Sousa, Wasp swarm algorithm for dynamic
  max-sat problems, in: International conference on adaptive and natural
  computing algorithms, Springer, 2007, pp. 350--357.

\bibitem{pan2012new}
W.-T. Pan, A new fruit fly optimization algorithm: taking the financial
  distress model as an example, Knowledge-Based Systems 26 (2012) 69--74.

\bibitem{hussain2019trade}
A.~Hussain, Y.~S. Muhammad, Trade-off between exploration and exploitation with
  genetic algorithm using a novel selection operator, Complex \& Intelligent
  Systems (2019) 1--14.

\bibitem{arani2013improved}
B.~O. Arani, P.~Mirzabeygi, M.~S. Panahi, An improved pso algorithm with a
  territorial diversity-preserving scheme and enhanced
  exploration--exploitation balance, Swarm and Evolutionary Computation 11
  (2013) 1--15.

\bibitem{gao2012global}
W.~Gao, S.~Liu, L.~Huang, A global best artificial bee colony algorithm for
  global optimization, Journal of Computational and Applied Mathematics
  236~(11) (2012) 2741--2753.

\bibitem{karaboga2009comparative}
D.~Karaboga, B.~Akay, A comparative study of artificial bee colony algorithm,
  Applied mathematics and computation 214~(1) (2009) 108--132.

\bibitem{yadav2018harmony}
N.~Yadav, A.~Yadav, J.~C. Bansal, K.~Deep, J.~H. Kim, Harmony Search and Nature
  Inspired Optimization Algorithms: Theory and Applications, ICHSA 2018, Vol.
  741, Springer, 2018.

\bibitem{olmstead1992cost}
K.~L. OLMSTEAD, R.~F. DENNO, Cost of shield defence for tortoise beetles
  (coleoptera: Chrysomelidae), Ecological Entomology 17~(3) (1992) 237--243.

\bibitem{olmstead1993effectiveness}
K.~L. Olmstead, R.~F. Denno, Effectiveness of tortoise beetle larval shields
  against different predator species, Ecology 74~(5) (1993) 1394--1405.

\bibitem{riley1870insects}
C.~Riley, Insects infesting the sweet potato. tortoise beetles (coleoptera,
  cassidae), Second Annual Report of the Noxious, Beneficial and other Insects
  of Missouri (1870) 56--64.

\bibitem{barrows1979life}
E.~M. Barrows, Life cycles, mating, and color change in tortoise beetles
  (coleoptera: Chrysomelidae: Cassidinae), The Coleopterists' Bulletin (1979)
  9--16.

\bibitem{vigneron2007switchable}
J.~P. Vigneron, J.~M. Pasteels, D.~M. Windsor, Z.~V{\'e}rtesy, M.~Rassart,
  T.~Seldrum, J.~Dumont, O.~Deparis, V.~Lousse, L.~P. Biro, et~al., Switchable
  reflector in the panamanian tortoise beetle charidotella egregia
  (chrysomelidae: Cassidinae), Physical Review E 76~(3) (2007) 031907.

\bibitem{lenau2008colours}
T.~Lenau, M.~Barfoed, Colours and metallic sheen in beetle shells—a
  biomimetic search for material structuring principles causing light
  interference, Advanced Engineering Materials 10~(4) (2008) 299--314.

\bibitem{vigneron2006spectral}
J.~P. Vigneron, M.~Rassart, C.~Vandenbem, V.~Lousse, O.~Deparis, L.~P.
  Bir{\'o}, D.~Dedouaire, A.~Cornet, P.~Defrance, Spectral filtering of visible
  light by the cuticle of metallic woodboring beetles and microfabrication of a
  matching bioinspired material, Physical Review E 73~(4) (2006) 041905.

\bibitem{capinera2015golden}
J.~Capinera, Golden tortoise beetle, charidotella (= metriona) bicolor
  (fabricius)(insecta: Coleoptera: Chrysomelidae) (2015).

\bibitem{yu2016improved}
F.~Yu, X.~Fu, H.~Li, G.~Dong, Improved roulette wheel selection-based genetic
  algorithm for tsp, in: 2016 International Conference on Network and
  Information Systems for Computers (ICNISC), IEEE, 2016, pp. 151--154.

\bibitem{liang2005novel}
J.-J. Liang, P.~N. Suganthan, K.~Deb, Novel composition test functions for
  numerical global optimization, in: Proceedings 2005 IEEE Swarm Intelligence
  Symposium, 2005. SIS 2005., IEEE, 2005, pp. 68--75.

\bibitem{poli2007particle}
R.~Poli, J.~Kennedy, T.~Blackwell, Particle swarm optimization, Swarm
  intelligence 1~(1) (2007) 33--57.

\bibitem{wilcoxon1945individual}
F.~Wilcoxon, Individual comparisons by ranking methods, Biometrics bulletin
  1~(6) (1945) 80--83.

\bibitem{coello2002theoretical}
C.~A.~C. Coello, Theoretical and numerical constraint-handling techniques used
  with evolutionary algorithms: a survey of the state of the art, Computer
  methods in applied mechanics and engineering 191~(11-12) (2002) 1245--1287.

\bibitem{pasandideh2015optimization}
S.~H.~R. Pasandideh, S.~T.~A. Niaki, A.~Gharaei, Optimization of a multiproduct
  economic production quantity problem with stochastic constraints using
  sequential quadratic programming, Knowledge-Based Systems 84 (2015) 98--107.

\bibitem{jalali2016optimizing}
S.~Jalali, M.~Seifbarghy, J.~Sadeghi, S.~Ahmadi, Optimizing a bi-objective
  reliable facility location problem with adapted stochastic measures using
  tuned-parameter multi-objective algorithms, Knowledge-Based Systems 95 (2016)
  45--57.

\bibitem{sayed2018new}
G.~I. Sayed, A.~Darwish, A.~E. Hassanien, A new chaotic multi-verse
  optimization algorithm for solving engineering optimization problems, Journal
  of Experimental \& Theoretical Artificial Intelligence 30~(2) (2018)
  293--317.

\bibitem{ragsdell1976optimal}
K.~Ragsdell, D.~Phillips, Optimal design of a class of welded structures using
  geometric programming.

\bibitem{he2004improved}
S.~He, E.~Prempain, Q.~Wu, An improved particle swarm optimizer for mechanical
  design optimization problems, Engineering optimization 36~(5) (2004)
  585--605.

\bibitem{gandomi2014interior}
A.~H. Gandomi, Interior search algorithm (isa): a novel approach for global
  optimization, ISA transactions 53~(4) (2014) 1168--1183.

\bibitem{sandgren1990nonlinear}
E.~Sandgren, Nonlinear integer and discrete programming in mechanical design
  optimization.

\end{thebibliography}







\end{document}